%% file: main.tex
\pdfoutput=1
\documentclass{article}

\PassOptionsToPackage{numbers,sort&compress}{natbib}

\usepackage[main, final]{neurips_2026}

\usepackage[utf8]{inputenc}
\usepackage[T1]{fontenc}
\usepackage{hyperref}
\usepackage{url}
\usepackage{booktabs}
\usepackage{amsthm,amsfonts,amsmath,amssymb,mathtools}
\usepackage{nicefrac}
\usepackage{microtype}
\usepackage{graphicx}
\usepackage{subcaption}
\usepackage{tabularx,multirow}
\usepackage{enumitem}  
\usepackage{prettyref}
\usepackage{algorithm}
\usepackage{algorithmic}
\newtheorem{proposition}{Proposition}

\usepackage{array}
\usepackage[table,dvipsnames]{xcolor}
\usepackage{tikz}
\usetikzlibrary{shadows,shadows.blur}
\usepackage[most]{tcolorbox}
\usepackage[capitalize,noabbrev]{cleveref}
\usepackage[textsize=tiny]{todonotes}

\usepackage{tabularx}
\usepackage{multirow}
\usepackage{booktabs}
\usepackage{caption}

\newcolumntype{C}{>{\centering\arraybackslash}X}
\usepackage{makecell}
\newcommand{\rescheck}{\ensuremath{\checkmark}}
\newcommand{\resx}{\ensuremath{\times}}

\newcommand{\resscore}[2][2.8em]{%
  \makebox[#1][c]{#2}%
  \makebox[2.8em][l]{}%
}
\newcommand{\resup}[3][2.8em]{%
  \makebox[#1][c]{#2}%
  \makebox[2.8em][l]{%
    {\scriptsize\,\textcolor{ForestGreen}{(+#3)}}}%
}
\newcommand{\resdown}[3][2.8em]{%
  \makebox[#1][c]{#2}%
  \makebox[2.8em][l]{%
    {\scriptsize\,\textcolor{red}{(-#3)}}}%
}
\newcommand{\ressame}[3][2.8em]{%
  \makebox[#1][c]{#2}%
  \makebox[2.8em][l]{%
    {\scriptsize\,\textcolor{gray}{(#3)}}}%
}

\newrefformat{alg}{Algorithm~\ref{#1}}
\newrefformat{tab}{Table~\ref{#1}}
\newrefformat{sec}{Section~\ref{#1}}
\newrefformat{fig}{Figure~\ref{#1}}
\newrefformat{eq}{Eq.~(\ref{#1})}
\newrefformat{app}{Appendix~\ref{#1}}
\newrefformat{ass}{Assumption~\ref{#1}}

\title{Separating Common and Unique Directions \\ for Model Merging}

\author{%
  Shiting Wang$^{1,2,3}$ \qquad
  Yingjie Zou$^{1,2,4}$ \qquad
  Ke Li$^{1,2,5}$ \thanks{Corresponding author}\\
  $^{1}$Earlham Institute, United Kingdom \\
$^{2}$Living Systems Institute, University of Exeter, United Kingdom \\
$^{3}$Department of Computer Science, University of Exeter, United Kingdom \\
$^{4}$Department of Engineering, University of Exeter, United Kingdom \\
$^{5}$School of Computing Sciences, University of East Anglia, United Kingdom \\
  \texttt{stwang0106@gmail.com},
  \texttt{yz1033@exeter.ac.uk},
  \texttt{ke.li@earlham.ac.uk} \\
}

\begin{document}

\maketitle

\begin{abstract}
Model merging combines multiple fine-tuned models without additional training, and arithmetic methods have emerged as a simple yet effective way to merge models without data. However, existing arithmetic methods treat task vectors in parameter space and apply a single element-wise rule to directions that can differ in mergeability. In a shared directional basis, we find a consistent two-regime structure: a low-dispersion common bulk shared across tasks and a high-dispersion task-dominated tail that carries disproportionate energy. These findings motivate \textbf{C}ommon--\textbf{U}nique \textbf{S}ubspace \textbf{D}ecomposition (CUSD), a data-free pre-merging method that separates task vectors into common and unique directional components, merges the common components with a downstream arithmetic operator, and retains unique residuals separately. Experiments across $11$ base models, four model families, and four arithmetic-based merging operators show that CUSD improves $42$ of $44$ model--operator combinations and achieves up to $22.5\%$ improvement over the corresponding baselines. Our code is publicly available at https://github.com/stwang666/CUSD
\end{abstract}

\input{sections/introduction}
\input{sections/related_work}
\input{sections/method}
\input{sections/experiment}

\input{sections/conclusion}

\clearpage

\begin{ack}
This work was supported by the UKRI Future Leaders Fellowship under grants MR/S017062/1 and MR/X011135/1; the National Natural Science Foundation of China under grants 62376056 and 62076056; Isambard AI Open Access (AIOA - 01 - 2026-06-04 - EvoRNA: Predictive Scaling of Evolution-Aware and Multimodal RNA Foundation Models for Crop Resilience); Modal Labs through a research grant of compute credits; the Royal Society Faraday Discovery Fellowship (FDF/S2/251014); BBSRC Transformative Research Technologies (UKRI1875); the Royal Society International Exchanges Award (IES/R3/243136); the Kan Tong Po Fellowship (KTP/R1/231017); the Amazon Research Award; and the Alan Turing Fellowship.
\end{ack}

\bibliographystyle{unsrtnat}
\bibliography{cusd}

\clearpage


\appendix

\section{Appendix}
\subsection{Additional Directional-Structure Evidence}

  \label{app:directional_structure}
\begin{figure}[htbp]
    \centering
    
    \includegraphics[width=0.8\textwidth]{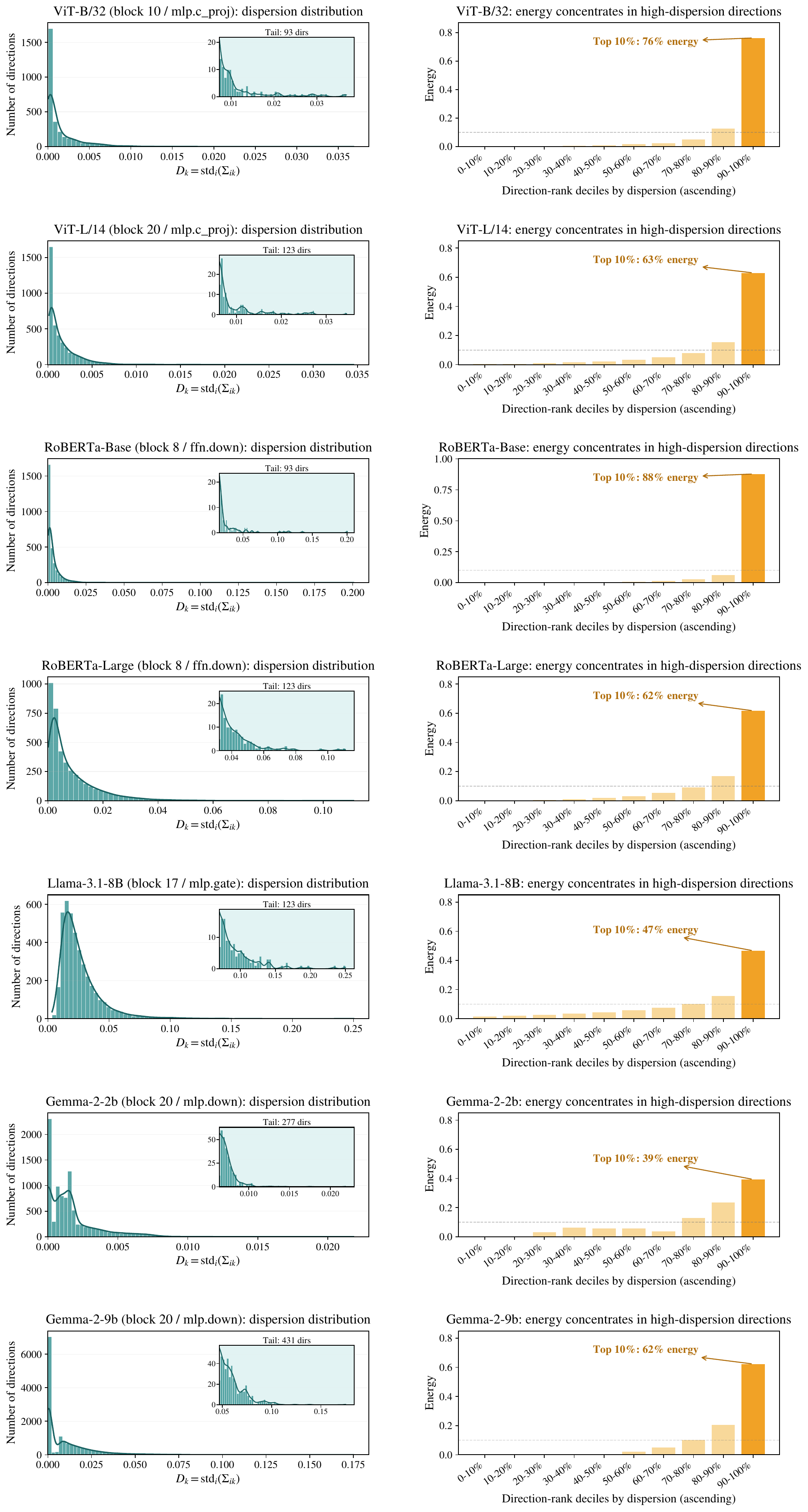}
    \caption{\textbf{Direction-level heterogeneity in task vectors.} Directional energy structure across all evaluated model families.}

    \label{fig:motivation_appendix}
  \end{figure}

Figure~\ref{fig:motivation_appendix} extends the motivation analysis in Figure~\ref{fig:motivation} to all evaluated model families, including ViT, RoBERTa, Llama, and Gemma. Each panel shows the task-vector directional energy distribution in the HO-GSVD-derived shared basis for a representative layer. Across architectures and task groups, the distributions consistently contain a low-dispersion bulk together with a high-dispersion tail of task-dominated directions. This recurring pattern supports the central premise of CUSD: arithmetic merging should first separate directions that are broadly common across tasks from directions whose energy is concentrated in one or a few task-specific models.

\subsection{HO-GSVD-Derived Shared-Basis Details}
\label{app:hogsvd_derivation}
 This appendix gives the regularized HO-GSVD-derived shared-basis construction used to compute $V$ and $\Sigma$ for each matrix-structured layer. The construction follows the pairwise Gram-quotient
  idea of HO-GSVD, with ridge regularization and symmetrization for neural-network task-vector matrices. The symmetrization step produces an orthonormal basis suitable for CUSD's projection step, so CUSD does not rely on an exact classical HO-GSVD factorization. Algorithm~\ref{alg:hogsvd} summarizes the implemented procedure.

  Let $A_1,\ldots,A_K$ denote the task-vector matrices for one matrix-structured layer, where $A_i\in\mathbb{R}^{m\times n}$ and all tasks share the same column dimension $n$. For a fixed $
  \pi>0$, define
  \begin{equation}
    G_i = A_i^\top A_i + \pi I_n,
    \qquad i=1,\ldots,K .
    \label{eq:regularized_gram}
  \end{equation}
  Each $G_i$ is symmetric positive definite, so all quotient terms below are well defined. We form the regularized pairwise quotient average
  \begin{equation}
    \widetilde{S}_\pi
    =
    \frac{1}{K(K-1)}
    \sum_{1\leq i<j\leq K}
    \left(
      G_j^{-1}G_i + G_i^{-1}G_j
    \right).
    \label{eq:regularized_pairwise_quotient}
  \end{equation}
  In practice, the products $G_j^{-1}G_i$ are computed by solving linear systems rather than explicitly forming matrix inverses. Since $\widetilde{S}_\pi$ is not guaranteed to be symmetric, CUSD
  uses
  \begin{equation}
    S_\pi
    =
    \frac{1}{2}
    \left(
      \widetilde{S}_\pi+\widetilde{S}_\pi^\top
    \right)
    \label{eq:symmetrized_pairwise_quotient}
  \end{equation}
  and computes the orthogonal eigendecomposition
  \begin{equation}
    S_\pi = V\Lambda V^\top,
    \qquad
    V^\top V = I_n .
    \label{eq:orthogonal_shared_basis}
  \end{equation}
  The columns of $V=[v_1,\ldots,v_n]$ are sorted by descending eigenvalue. The symmetrization step preserves the pairwise quotient signal while producing an orthonormal basis for the projection
  step in CUSD.

  For each task $i$ and shared direction $v_k$, CUSD records the raw directional energy
  \begin{equation}
    \sigma_{ik}
    =
    \|A_i v_k\|_2,
    \qquad
    \Sigma=[\sigma_{ik}]\in\mathbb{R}^{K\times n}.
    \label{eq:directional_energy}
  \end{equation}
  These raw directional energies form the matrix $\Sigma$ used by CUSD for dispersion scoring.

  \begin{algorithm}[h]
  \caption{Regularized HO-GSVD-Derived Shared-Basis Construction}
  \label{alg:hogsvd}
  \begin{algorithmic}[1]
  \REQUIRE $A_1,\ldots,A_K$ with $A_i\in\mathbb{R}^{m_i\times n}$; $K\geq 2$; $\pi=10^{-3}$
  \ENSURE $V\in\mathbb{R}^{n\times n}$ (orthogonal shared basis); $\Sigma\in\mathbb{R}^{K\times n}$ (directional-energy magnitudes)
  \STATE \textbf{Regularized Grams:} $G_i\gets A_i^\top A_i+\pi I_n$ for $i=1,\ldots,K$
  \STATE \textbf{Pairwise quotient average:}
  $\displaystyle
  \widetilde{S}_\pi
  \gets
  \frac{1}{K(K-1)}
  \sum_{1\leq i<j\leq K}
  \left(
  \operatorname{solve}(G_j,G_i)+\operatorname{solve}(G_i,G_j)
  \right)$
  \STATE \textbf{Symmetrization:} $S_\pi\gets\frac{1}{2}\left(\widetilde{S}_\pi+\widetilde{S}_\pi^\top\right)$
  \STATE \textbf{Orthogonal eigendecomposition:} $S_\pi=V\Lambda V^\top$; sort columns of $V$ by descending eigenvalue
  \STATE \textbf{Magnitude extraction:} $\sigma_{ik}\gets\|A_i v_k\|_2$ for all $i\in[K]$ and $k\in[n]$
  \RETURN $V,\;\Sigma=[\sigma_{ik}]$
	  \end{algorithmic}
	  \end{algorithm}

\begin{proposition}[Well-defined shared-basis construction]
\label{prop:shared_basis}
For any $\pi>0$, the CUSD shared-basis construction in Algorithm~\ref{alg:hogsvd} is well defined and returns an orthonormal basis $V$.
\end{proposition}

\begin{proof}
For each task matrix $A_i$, the matrix $G_i=A_i^\top A_i+\pi I_n$ is symmetric positive definite because $\pi>0$. Hence all linear systems involving $G_i$ in Eq.~\ref{eq:regularized_pairwise_quotient} are well defined. The symmetrized quotient matrix $S_\pi=(\widetilde{S}_\pi+\widetilde{S}_\pi^\top)/2$ is real symmetric. By the spectral theorem, it admits an eigendecomposition $S_\pi=V\Lambda V^\top$ with $V^\top V=I_n$. Thus the construction returns an orthonormal shared basis.
\end{proof}

The orthogonality of $V$ is the key property used in the lossless decomposition proof in Appendix~\ref{app:lossless}.

\subsection{KDE Classification Details}
  \label{app:kde}

  CUSD selects the common--unique threshold from the empirical distribution of dispersion scores using kernel density estimation (KDE)~\citep{scott2015multivariate}. This
  provides a layer-adaptive threshold without manually fixing a global cutoff.

  \paragraph{Density Estimation Framework}
  Given a set of $n$ dispersion scores $\mathcal{D}=\{d_k\}_{k=1}^n$, we construct the density estimate $\hat{f}(d)$ using a Gaussian kernel:
  \begin{equation}
      \hat{f}(d) = \frac{1}{nh} \sum_{k=1}^n \frac{1}{\sqrt{2\pi}} \exp\left( -\frac{(d - d_k)^2}{2h^2} \right).
  \end{equation}
  The bandwidth $h$ is set by Scott's rule for univariate data:
  \begin{equation}
      h = \hat{\sigma} n^{-1/5},
  \end{equation}
  where $\hat{\sigma}$ denotes the sample standard deviation of $\mathcal{D}$. We evaluate $\hat{f}(d)$ over a uniform grid of $N=1{,}000$ points covering $[\min_k d_k, \max_k d_k]$.

  \paragraph{Threshold Selection and Decision Rule}
	  The algorithm identifies candidate thresholds by locating local minima of the estimated density. Let $\mathcal{V}$ be the set of grid points corresponding to these local minima. If $
	  \mathcal{V}$ is nonempty, CUSD selects the valley closest to the midpoint of the score range:
	  \begin{equation}
	      \delta = \arg\min_{v \in \mathcal{V}} \left| v - \frac{\min_k d_k + \max_k d_k}{2} \right|.
	  \end{equation}
	  The final partition is
	  \[
	      \mathcal{C}=\{k:d_k<\delta\},
	      \qquad
	      \mathcal{U}=\{k:d_k\geq\delta\}.
	  \]
  If no valid valley is found, CUSD applies a conservative no-valley policy and assigns all directions in that layer to the common set $\mathcal{C}$.

\subsection{Lossless Orthogonal Decomposition}
\label{app:lossless}

\begin{proposition}
\label{prop:lossless}
Let $P_\mathcal{C} = V_\mathcal{C} V_\mathcal{C}^\top$ be the orthogonal projector onto the common directions, and let $P_\mathcal{U} = I - P_\mathcal{C}$. Set $C_i = A_i P_\mathcal{C}$ and $R_i = A_i P_\mathcal{U}$. Then for every task $i$:
(i)~$C_i + R_i = A_i$; \;
(ii)~$\langle C_i, R_i \rangle_F = 0$.
\end{proposition}

\begin{proof}
The claim uses only the fact that the CUSD basis $V$ is orthonormal.
Let $V_\mathcal{C}$ contain the columns of $V$ indexed by $\mathcal{C}$.
Since $V^\top V=I$, the selected columns also satisfy
$V_\mathcal{C}^\top V_\mathcal{C}=I$. Hence
\[
P_\mathcal{C}^\top
=
(V_\mathcal{C}V_\mathcal{C}^\top)^\top
=
P_\mathcal{C},
\qquad
P_\mathcal{C}^2
=
V_\mathcal{C}(V_\mathcal{C}^\top V_\mathcal{C})V_\mathcal{C}^\top
=
P_\mathcal{C}.
\]
Thus $P_\mathcal{C}$ is the orthogonal projector onto the span of the
common directions. Since $P_\mathcal{U}=I-P_\mathcal{C}$, we also have
\[
P_\mathcal{U}+P_\mathcal{C}=I,
\qquad
P_\mathcal{U}P_\mathcal{C}
=(I-P_\mathcal{C})P_\mathcal{C}
=P_\mathcal{C}-P_\mathcal{C}^2
=0.
\]
Therefore,
\[
C_i+R_i
=
A_iP_\mathcal{C}+A_iP_\mathcal{U}
=
A_i(P_\mathcal{C}+P_\mathcal{U})
=
A_i.
\]
For orthogonality under the Frobenius inner product,
\[
\langle C_i,R_i\rangle_F
=
\operatorname{tr}(C_i^\top R_i)
=
\operatorname{tr}\!\left(P_\mathcal{C}A_i^\top A_iP_\mathcal{U}\right).
\]
Using cyclic invariance of the trace,
\[
\operatorname{tr}\!\left(P_\mathcal{C}A_i^\top A_iP_\mathcal{U}\right)
=
\operatorname{tr}\!\left(P_\mathcal{U}P_\mathcal{C}A_i^\top A_i\right)
=0,
\]
because $P_\mathcal{U}P_\mathcal{C}=0$. Hence
$\langle C_i,R_i\rangle_F=0$ for every task $i$.
\end{proof}

\subsection{Experimental Setup}
\label{app:setup}

\paragraph{Models.}
We evaluate CUSD on 11 base models from four model families (ViT, RoBERTa, Llama, and Gemma), covering vision encoders, text encoders, and decoder-only LLMs. For vision experiments, we use CLIP ViT-B/32, ViT-B/16, and ViT-L/14 models~\citep{radford2021learning,dosovitskiy2020image}. For Natural Language Understanding (NLU) experiments, we use RoBERTa-Base and RoBERTa-Large~\citep{liu2019roberta}. For generative LLM experiments, we use the MergeBench fine-tuned models~\citep{he2025mergebench} based on Llama-3.2-3B, Llama-3.1-8B, Llama-3.2-3B-Instruct, Llama-3.1-8B-Instruct, Gemma-2-2b, and Gemma-2-9b~\citep{grattafiori2024llama,team2024gemma}.

\paragraph{Datasets.}
For vision tasks, we use eight image classification datasets following prior work~\citep{yang2023adamerging,cheng2025whoever}: Cars, DTD, EuroSAT, GTSRB, MNIST, RESISC45, SUN397, and SVHN. For NLU tasks, we follow prior model-merging evaluations~\citep{cheng2025whoever} and use the eight GLUE tasks~\citep{wang2018glue}: CoLA, SST-2, MRPC, STS-B, QQP, MNLI, QNLI, and RTE. For generative language tasks, we follow MergeBench~\citep{he2025mergebench} and evaluate code generation with Humaneval+ and MBPP+, mathematical reasoning with GSM8k, and multilingual understanding with M-MMLU, M-ARC, and M-Hellaswag.

\paragraph{Baselines.}
We evaluate CUSD with four representative data-free arithmetic-based merging methods: Task Arithmetic~\citep{ilharco2022editing}, TIES-Merging~\citep{yadav2023ties}, DARE~\citep{yu2024language}, and Consensus Merging~\citep{wang2024localizing}. For DARE and Consensus Merging, we use their Task Arithmetic instantiations. For each method, we report the plain merged model and the corresponding CUSD-preprocessed variant. We also include TSV-Merge~\citep{gargiulo2025task} and Iso-Merging~\citep{marczak2025no} for comparison with subspace-based merging methods. Fine-tuned task-specific models are included as reference upper bounds where available.

\paragraph{Implementation details.}
CUSD operates on matrix-structured linear-layer task vectors.
We use absolute scoring for encoder models (ViT and RoBERTa)
and relative scoring for decoder LLMs.
Section~\ref{sec:ablation} compares the dispersion scoring modes
and analyzes sensitivity to the unique-residual coefficient
$\beta$. Figure~\ref{fig:beta_sweep} provides empirical guidance
on suitable $\beta$ ranges for the evaluated model families.
Given fixed hyperparameters, CUSD uses only model weights
during decomposition and merging.
Baseline hyperparameters follow the original papers or
MergeBench defaults and are kept identical between plain and
CUSD-preprocessed runs.

\subsection{Main Results Details}
\label{app:main_results}

\paragraph{Metrics.}
We report normalized scores relative to the corresponding fine-tuned task-specific references. For ViT and RoBERTa results in Table~\ref{tab:overall_performance}, task scores are first averaged in their original metric scale and the resulting average is normalized by the corresponding fine-tuned average. For LLMs, following ~\citep{he2025mergebench}, each benchmark score is normalized by the corresponding fine-tuned benchmark score before averaging.

\paragraph{Results on vision tasks.}
Tables~\ref{tab:vit_l14}--\ref{tab:vit_b32} provide task-level results for the three ViT models used in the vision experiments. Each table reports the plain arithmetic merge and the corresponding CUSD merge on the eight image classification tasks. CUSD improves all four downstream operators on the three CLIP ViT backbones. The gains are largest on ViT-B/32 and ViT-B/16, where the average improvement across operators is about $12$ normalized points, and remain positive on ViT-L/14, with an average gain of about $5$ normalized points. The task-level results show a clear trade-off. For example, on ViT-L/14 with TIES-Merging, CUSD slightly reduces the already strong MNIST accuracy by $0.1$
points, but improves EuroSAT, RESISC45, and DTD by $10.7$, $8.5$, and $9.2$ points, respectively, raising the average task accuracy by $5.9$ points. This pattern is consistent with CUSD reducing the influence of task-dominated directions that would otherwise bias the arithmetic merge, thereby improving the overall multi-task balance. TIES-Merging with CUSD achieves the best normalized average on all three ViT models. Detailed task-level results are provided in Appendix Tables~\ref{tab:vit_l14}--\ref{tab:vit_b32}.

\begin{table}[H]
\caption{Multi-task performance of merged ViT-L/14 models on eight tasks.}
\label{tab:vit_l14}
\centering
\small
\setlength{\tabcolsep}{2.5pt}
\resizebox{\linewidth}{!}{%
\begin{tabular}{lc cccccccc c}
\toprule
\textbf{Method} & with \textbf{CUSD} & \textbf{Cars} & \textbf{DTD} & \textbf{EuroSAT} & \textbf{GTSRB} & \textbf{MNIST} & \textbf{RESISC45} & \textbf{SUN397} & \textbf{SVHN} & \textbf{Avg.} \\
\midrule
Fine-tuned & --- & \resscore{92.5} & \resscore{84.2} & \resscore{91.0} & \resscore{99.2} & \resscore{99.7} & \resscore{99.7} & \resscore{83.6} & \resscore{98.1} & \resscore{93.5} \\
\midrule
\multirow{2}{*}{Task Arithmetic}
 & \resx & \resscore{80.5} & \resscore{64.6} & \resscore{75.6} & \resscore{86.7} & \resscore{99.0} & \resscore{86.1} & \resscore{72.9} & \resscore{89.7} & \resscore{81.9} \\
 & \rescheck & \resup{87.2}{6.7} & \resup{71.9}{7.2} & \resup{87.7}{12.0} & \resup{92.6}{6.0} & \resdown{98.7}{0.3} & \resup{94.5}{8.4} & \resup{76.1}{3.2} & \resdown{87.2}{2.5} & \resup{\textbf{87.0}}{5.1} \\
\midrule
\multirow{2}{*}{TIES-Merging}
 & \resx & \resscore{81.1} & \resscore{65.3} & \resscore{77.8} & \resscore{87.5} & \resscore{99.1} & \resscore{87.2} & \resscore{73.3} & \resscore{90.5} & \resscore{82.7} \\
 & \rescheck & \resup{87.7}{6.6} & \resup{74.5}{9.2} & \resup{88.4}{10.7} & \resup{95.0}{7.5} & \resdown{99.0}{0.1} & \resup{95.7}{8.5} & \resup{77.2}{3.9} & \resup{91.2}{0.8} & \resup{\textbf{88.6}}{5.9} \\
\midrule
\multirow{2}{*}{DARE}
 & \resx & \resscore{80.4} & \resscore{64.0} & \resscore{75.6} & \resscore{86.7} & \resscore{99.0} & \resscore{86.1} & \resscore{72.8} & \resscore{89.7} & \resscore{81.8} \\
 & \rescheck & \resup{87.1}{6.7} & \resup{71.6}{7.6} & \resup{87.3}{11.8} & \resup{92.7}{6.0} & \resdown{98.7}{0.3} & \resup{94.4}{8.3} & \resup{76.0}{3.3} & \resdown{87.2}{2.6} & \resup{\textbf{86.9}}{5.1} \\
\midrule
\multirow{2}{*}{Consensus}
 & \resx & \resscore{80.8} & \resscore{65.0} & \resscore{76.7} & \resscore{87.0} & \resscore{99.0} & \resscore{86.6} & \resscore{73.0} & \resscore{89.7} & \resscore{82.2} \\
 & \rescheck & \resup{87.2}{6.4} & \resup{71.4}{6.4} & \resup{87.6}{10.9} & \resup{92.4}{5.4} & \resdown{98.7}{0.4} & \resup{94.3}{7.7} & \resup{75.9}{2.9} & \resdown{86.8}{2.9} & \resup{\textbf{86.8}}{4.6} \\
\bottomrule
\end{tabular}
}
\end{table}

\begin{table}[H]
\caption{Multi-task performance of merged ViT-B/16 models on eight tasks.}
\label{tab:vit_b16}
\centering
\small
\setlength{\tabcolsep}{2.5pt}
\resizebox{\linewidth}{!}{%
\begin{tabular}{lc cccccccc c}
\toprule
\textbf{Method} & with \textbf{CUSD} & \textbf{Cars} & \textbf{DTD} & \textbf{EuroSAT} & \textbf{GTSRB} & \textbf{MNIST} & \textbf{RESISC45} & \textbf{SUN397} & \textbf{SVHN} & \textbf{Avg.} \\
\midrule
Fine-tuned & --- & \resscore{86.5} & \resscore{82.1} & \resscore{90.2} & \resscore{99.2} & \resscore{99.8} & \resscore{99.7} & \resscore{80.6} & \resscore{97.9} & \resscore{92.0} \\
\midrule
\multirow{2}{*}{Task Arithmetic}
 & \resx & \resscore{59.0} & \resscore{51.4} & \resscore{64.9} & \resscore{76.3} & \resscore{98.5} & \resscore{71.3} & \resscore{54.3} & \resscore{89.2} & \resscore{70.6} \\
 & \rescheck & \resup{75.6}{16.5} & \resup{62.1}{10.7} & \resup{85.7}{20.8} & \resup{87.1}{10.9} & \resdown{97.9}{0.5} & \resup{89.2}{17.9} & \resup{71.6}{17.3} & \resdown{84.7}{4.5} & \resup{\textbf{81.7}}{11.1} \\
\midrule
\multirow{2}{*}{TIES-Merging}
 & \resx & \resscore{60.7} & \resscore{52.3} & \resscore{68.5} & \resscore{77.1} & \resscore{98.8} & \resscore{73.4} & \resscore{56.6} & \resscore{90.2} & \resscore{72.2} \\
 & \rescheck & \resup{74.7}{14.0} & \resup{65.2}{12.9} & \resup{86.8}{18.3} & \resup{90.3}{13.3} & \resdown{98.4}{0.4} & \resup{90.2}{16.8} & \resup{70.7}{14.1} & \resdown{88.4}{1.9} & \resup{\textbf{83.1}}{10.9} \\
\midrule
\multirow{2}{*}{DARE}
 & \resx & \resscore{59.2} & \resscore{51.3} & \resscore{64.8} & \resscore{75.6} & \resscore{98.4} & \resscore{71.3} & \resscore{54.2} & \resscore{89.2} & \resscore{70.5} \\
 & \rescheck & \resup{75.3}{16.1} & \resup{62.1}{10.8} & \resup{85.6}{20.8} & \resup{87.1}{11.4} & \resdown{97.8}{0.6} & \resup{89.2}{18.0} & \resup{71.6}{17.4} & \resdown{84.9}{4.3} & \resup{\textbf{81.7}}{11.2} \\
\midrule
\multirow{2}{*}{Consensus}
 & \resx & \resscore{60.8} & \resscore{52.1} & \resscore{67.1} & \resscore{77.1} & \resscore{98.6} & \resscore{72.9} & \resscore{56.2} & \resscore{89.5} & \resscore{71.8} \\
 & \rescheck & \resup{75.7}{15.0} & \resup{62.1}{9.9} & \resup{85.7}{18.5} & \resup{86.7}{9.6} & \resdown{97.9}{0.7} & \resup{89.5}{16.6} & \resup{71.7}{15.5} & \resdown{84.4}{5.1} & \resup{\textbf{81.7}}{9.9} \\
\bottomrule
\end{tabular}
}
\end{table}

\begin{table}[H]
\caption{Multi-task performance of merged ViT-B/32 models on eight tasks.}
\label{tab:vit_b32}
\centering
\small
\setlength{\tabcolsep}{2.5pt}
\resizebox{\linewidth}{!}{%
\begin{tabular}{lc cccccccc c}
\toprule
\textbf{Method} & with \textbf{CUSD} & \textbf{Cars} & \textbf{DTD} & \textbf{EuroSAT} & \textbf{GTSRB} & \textbf{MNIST} & \textbf{RESISC45} & \textbf{SUN397} & \textbf{SVHN} & \textbf{Avg.} \\
\midrule
Fine-tuned & --- & \resscore{78.0} & \resscore{79.4} & \resscore{88.9} & \resscore{98.7} & \resscore{99.7} & \resscore{99.7} & \resscore{78.3} & \resscore{97.5} & \resscore{90.0} \\
\midrule
\multirow{2}{*}{Task Arithmetic}
 & \resx & \resscore{46.4} & \resscore{48.5} & \resscore{64.2} & \resscore{71.4} & \resscore{97.2} & \resscore{64.1} & \resscore{46.2} & \resscore{79.3} & \resscore{64.7} \\
 & \rescheck & \resup{65.1}{18.8} & \resup{60.4}{12.0} & \resup{81.9}{17.7} & \resup{79.8}{8.5} & \resdown{96.5}{0.7} & \resup{83.4}{19.3} & \resup{65.5}{19.4} & \ressame{79.3}{0.0} & \resup{\textbf{76.5}}{11.9} \\
\midrule
\multirow{2}{*}{TIES-Merging}
 & \resx & \resscore{48.8} & \resscore{49.7} & \resscore{66.9} & \resscore{72.8} & \resscore{97.7} & \resscore{66.7} & \resscore{49.0} & \resscore{81.3} & \resscore{66.6} \\
 & \rescheck & \resup{65.6}{16.8} & \resup{63.7}{14.0} & \resup{83.8}{16.9} & \resup{83.7}{10.9} & \resdown{96.8}{0.9} & \resup{87.1}{20.4} & \resup{66.1}{17.1} & \resdown{79.8}{1.5} & \resup{\textbf{78.3}}{11.7} \\
\midrule
\multirow{2}{*}{DARE}
 & \resx & \resscore{45.9} & \resscore{48.6} & \resscore{64.2} & \resscore{71.4} & \resscore{97.2} & \resscore{63.7} & \resscore{45.9} & \resscore{79.1} & \resscore{64.5} \\
 & \rescheck & \resup{64.6}{18.7} & \resup{60.3}{11.7} & \resup{81.8}{17.6} & \resup{79.9}{8.5} & \resdown{96.5}{0.7} & \resup{83.2}{19.5} & \resup{65.5}{19.6} & \resdown{79.1}{0.1} & \resup{\textbf{76.4}}{11.9} \\
\midrule
\multirow{2}{*}{Consensus}
 & \resx & \resscore{48.2} & \resscore{49.6} & \resscore{65.7} & \resscore{72.2} & \resscore{97.3} & \resscore{65.7} & \resscore{48.4} & \resscore{80.0} & \resscore{65.9} \\
 & \rescheck & \resup{65.3}{17.1} & \resup{60.5}{10.9} & \resup{82.0}{16.3} & \resup{79.6}{7.4} & \resdown{96.4}{0.9} & \resup{83.5}{17.8} & \resup{65.8}{17.4} & \resdown{79.2}{0.8} & \resup{\textbf{76.5}}{10.7} \\
\bottomrule
\end{tabular}
}
\end{table}

\paragraph{Results on NLU tasks.}
Tables~\ref{tab:roberta_base} and~\ref{tab:roberta_large} provide task-level GLUE results for RoBERTa-Base and RoBERTa-Large. Each table reports the plain merge and the corresponding CUSD merge for all four downstream arithmetic operators.
CUSD also improves every merging operator on both RoBERTa-Base and RoBERTa-Large. The effect is especially strong for operators whose plain merges suffer from severe degradation, such as TIES-Merging on RoBERTa-Large, where the normalized average increases from $47.3$ to $69.8$. A striking example is STS-B, the semantic textual similarity task: all plain merges yield negative normalized scores, whereas CUSD turns them positive across every operator in both RoBERTa-Base and RoBERTa-Large. These results indicate that common--unique separation is not limited to vision encoders and can also recover lost performance in discriminative language models. Detailed GLUE results are reported in Appendix Tables~\ref{tab:roberta_large} and \ref{tab:roberta_base}.

\begin{table}[H]
\caption{ Multi-task performance of merged RoBERTa-Base models on the 8-task GLUE benchmark.}
\label{tab:roberta_base}
\centering
\small
\setlength{\tabcolsep}{2.5pt}
\resizebox{\linewidth}{!}{%
\begin{tabular}{lc cccccccc c}
\toprule
\textbf{Method} & with \textbf{CUSD} & \textbf{CoLA} & \textbf{MNLI} & \textbf{MRPC} & \textbf{QNLI} & \textbf{QQP} & \textbf{RTE} & \textbf{SST-2} & \textbf{STS-B} & \textbf{Avg.} \\
\midrule
Fine-tuned & --- & \resscore{55.1} & \resscore{87.5} & \resscore{88.0} & \resscore{92.4} & \resscore{89.4} & \resscore{67.5} & \resscore{93.8} & \resscore{88.3} & \resscore{82.8} \\
\midrule
\multirow{2}{*}{Task Arithmetic}
 & \resx & \resscore{0.0} & \resscore{64.6} & \resscore{79.5} & \resscore{75.5} & \resscore{71.3} & \resscore{49.1} & \resscore{71.1} & \resscore{-56.8} & \resscore{44.3} \\
 & \rescheck & \resup{6.6}{6.6} & \resdown{63.6}{1.0} & \resdown{78.2}{1.3} & \resdown{74.1}{1.4} & \resdown{37.6}{33.7} & \resup{50.9}{1.8} & \resup{87.7}{16.6} & \resup{55.9}{112.6} & \resup{\textbf{56.8}}{12.5} \\
\midrule
\multirow{2}{*}{TIES-Merging}
 & \resx & \resscore{0.0} & \resscore{64.9} & \resscore{79.0} & \resscore{73.2} & \resscore{76.4} & \resscore{46.2} & \resscore{65.9} & \resscore{-67.2} & \resscore{42.3} \\
 & \rescheck & \resup{6.6}{6.6} & \resdown{62.1}{2.8} & \resdown{77.9}{1.1} & \resdown{63.6}{9.6} & \resdown{36.6}{39.8} & \resup{53.4}{7.2} & \resup{90.3}{24.3} & \resup{68.4}{135.6} & \resup{\textbf{57.4}}{15.1} \\
\midrule
\multirow{2}{*}{DARE}
 & \resx & \resscore{4.6} & \resscore{64.6} & \resscore{79.4} & \resscore{75.6} & \resscore{71.7} & \resscore{48.7} & \resscore{71.3} & \resscore{-58.6} & \resscore{44.7} \\
 & \rescheck & \resup{6.6}{1.9} & \resdown{63.0}{1.6} & \resdown{78.1}{1.3} & \resdown{73.6}{2.1} & \resdown{37.4}{34.3} & \resup{50.9}{2.2} & \resup{87.7}{16.4} & \resup{54.8}{113.3} & \resup{\textbf{56.5}}{11.8} \\
\midrule
\multirow{2}{*}{Consensus}
 & \resx & \resscore{4.6} & \resscore{66.9} & \resscore{78.0} & \resscore{78.6} & \resscore{71.8} & \resscore{46.6} & \resscore{75.5} & \resscore{-56.3} & \resscore{45.7} \\
 & \rescheck & \resup{6.6}{1.9} & \resdown{59.4}{7.5} & \resdown{77.3}{0.7} & \resdown{66.6}{12.0} & \resdown{33.3}{38.5} & \resup{53.1}{6.5} & \resup{87.7}{12.3} & \resup{61.6}{117.9} & \resup{\textbf{55.7}}{10.0} \\
\bottomrule
\end{tabular}
}
\end{table}

\begin{table}[H]
\caption{ Multi-task performance of merged RoBERTa-Large models on the 8-task GLUE benchmark.}
\label{tab:roberta_large}
\centering
\small
\setlength{\tabcolsep}{2.5pt}
\resizebox{\linewidth}{!}{%
\begin{tabular}{lc cccccccc c}
\toprule
\textbf{Method} & with \textbf{CUSD} & \textbf{CoLA} & \textbf{MNLI} & \textbf{MRPC} & \textbf{QNLI} & \textbf{QQP} & \textbf{RTE} & \textbf{SST-2} & \textbf{STS-B} & \textbf{Avg.} \\
\midrule
Fine-tuned & --- & \resscore{59.1} & \resscore{90.5} & \resscore{89.0} & \resscore{94.5} & \resscore{90.5} & \resscore{70.0} & \resscore{96.6} & \resscore{91.7} & \resscore{85.2} \\
\midrule
\multirow{2}{*}{Task Arithmetic}
 & \resx & \resscore{0.2} & \resscore{75.7} & \resscore{48.6} & \resscore{69.5} & \resscore{82.1} & \resscore{58.8} & \resscore{85.1} & \resscore{-22.4} & \resscore{49.7} \\
 & \rescheck & \resup{9.1}{8.8} & \resdown{72.2}{3.4} & \resdown{27.9}{20.8} & \resup{83.5}{14.0} & \resdown{59.5}{22.7} & \resup{70.8}{11.9} & \resup{92.7}{7.6} & \resup{62.0}{84.3} & \resup{\textbf{59.7}}{10.0} \\
\midrule
\multirow{2}{*}{TIES-Merging}
 & \resx & \resscore{-1.4} & \resscore{71.0} & \resscore{56.3} & \resscore{53.3} & \resscore{84.6} & \resscore{54.5} & \resscore{72.1} & \resscore{-67.9} & \resscore{40.3} \\
 & \rescheck & \resup{5.7}{7.1} & \resup{74.5}{3.5} & \resdown{29.1}{27.2} & \resup{85.3}{32.0} & \resdown{58.6}{26.0} & \resup{70.0}{15.5} & \resup{92.6}{20.4} & \resup{60.3}{128.2} & \resup{\textbf{59.5}}{19.2} \\
\midrule
\multirow{2}{*}{DARE}
 & \resx & \resscore{2.4} & \resscore{75.6} & \resscore{48.8} & \resscore{70.2} & \resscore{82.1} & \resscore{59.2} & \resscore{85.7} & \resscore{-23.2} & \resscore{50.1} \\
 & \rescheck & \resup{9.6}{7.2} & \resdown{71.7}{3.9} & \resdown{26.9}{21.9} & \resup{82.9}{12.7} & \resdown{59.1}{23.0} & \resup{69.7}{10.5} & \resup{92.4}{6.8} & \resup{61.3}{84.5} & \resup{\textbf{59.2}}{9.1} \\
\midrule
\multirow{2}{*}{Consensus}
 & \resx & \resscore{2.4} & \resscore{76.3} & \resscore{40.1} & \resscore{76.8} & \resscore{80.3} & \resscore{57.8} & \resscore{89.1} & \resscore{-0.9} & \resscore{52.7} \\
 & \rescheck & \resup{9.9}{7.5} & \resdown{71.7}{4.6} & \resdown{27.9}{12.2} & \resup{83.4}{6.6} & \resdown{59.9}{20.4} & \resup{70.8}{13.0} & \resup{92.7}{3.6} & \resup{61.1}{62.1} & \resup{\textbf{59.7}}{6.9} \\
\bottomrule
\end{tabular}
}
\end{table}

\paragraph{Results on generative LLMs.}
Tables~\ref{tab:llm_llama32_3b}--\ref{tab:llm_gemma2_9b} provide task-level results for the six LLMs: Llama-3.2-3B, Llama-3.1-8B, Llama-3.2-3B-Instruct, Llama-3.1-8B-Instruct, Gemma-2-2B, and Gemma-2-9B. Scores are normalized by the corresponding fine-tuned task-specific models before averaging. On generative LLMs, CUSD improves most settings, although the margins are more model-dependent than on encoder models. The strongest gains appear on Llama-3.2-3B-Instruct, Llama-3.1-8B-Instruct, and the
  Gemma models. Llama-3.1-8B is the least responsive case: plain arithmetic merges already reach normalized averages around $94$, leaving little headroom, and the only negative change is a small
  $0.3$-point decrease for DARE. In the evaluated LLM settings, partial unique-residual scaling is often beneficial; this behavior is examined in the $\beta$ ablation (Section~\ref{sec:ablation}) and the analysis of unique directions across models (Section~\ref{sec:analysis}).  
  Full task-level LLM results are provided in the Appendix Tables~\ref{tab:llm_llama32_3b}--\ref{tab:llm_gemma2_9b}.

\begin{table}[H]
\caption{Performance of merged Llama-3.2-3B models. The task-specific checkpoints are fine-tuned for coding, math, and multilingual tasks; normalized scores and their average are reported.}
\label{tab:llm_llama32_3b}
\centering
\small
\setlength{\tabcolsep}{2.5pt}
\resizebox{\linewidth}{!}{%
\begin{tabular}{lc cccccc c}
\toprule
\textbf{Method} & with \textbf{CUSD} & \textbf{Humaneval+} & \textbf{MBPP+} & \textbf{GSM8k} & \textbf{M-MMLU} & \textbf{M-ARC} & \textbf{M-HellaSwag} & \textbf{Avg.} \\
\midrule
Fine-tuned & --- & \resscore{100.0} & \resscore{100.0} & \resscore{100.0} & \resscore{100.0} & \resscore{100.0} & \resscore{100.0} & \resscore{100.0} \\
\midrule
\multirow{2}{*}{Task Arithmetic}
 & \resx & \resscore{45.7} & \resscore{59.4} & \resscore{39.2} & \resscore{87.0} & \resscore{93.4} & \resscore{93.0} & \resscore{\textbf{69.6}} \\
 & \rescheck & \ressame{45.7}{0.0} & \resdown{52.8}{6.6} & \resup{45.8}{6.6} & \resdown{86.0}{1.0} & \resup{94.0}{0.6} & \resup{93.2}{0.2} & \ressame{\textbf{69.6}}{0.0} \\
\midrule
\multirow{2}{*}{TIES-Merging}
 & \resx & \resscore{46.1} & \resscore{58.5} & \resscore{37.9} & \resscore{87.2} & \resscore{93.1} & \resscore{92.6} & \resscore{69.2} \\
 & \rescheck & \resdown{45.2}{0.9} & \resdown{53.3}{5.2} & \resup{47.1}{9.2} & \resdown{85.8}{1.4} & \resup{94.0}{0.9} & \resup{93.0}{0.4} & \resup{\textbf{69.7}}{0.5} \\
\midrule
\multirow{2}{*}{DARE}
 & \resx & \resscore{47.3} & \resscore{53.8} & \resscore{38.8} & \resscore{87.2} & \resscore{93.3} & \resscore{93.0} & \resscore{68.9} \\
 & \rescheck & \resup{47.9}{0.6} & \resup{57.1}{3.3} & \resup{42.4}{3.6} & \resdown{86.8}{0.4} & \resup{94.2}{0.9} & \resup{93.2}{0.2} & \resup{\textbf{70.3}}{1.4} \\
\midrule
\multirow{2}{*}{Consensus}
 & \resx & \resscore{47.9} & \resscore{59.9} & \resscore{24.4} & \resscore{87.8} & \resscore{91.5} & \resscore{92.8} & \resscore{67.4} \\
 & \rescheck & \resup{52.1}{4.2} & \resdown{51.9}{8.0} & \resup{47.1}{22.7} & \resdown{85.7}{2.1} & \resup{93.9}{2.4} & \resup{93.1}{0.3} & \resup{\textbf{70.6}}{3.2} \\
\bottomrule
\end{tabular}
}
\end{table}

\begin{table}[H]
\caption{Performance of merged Llama-3.1-8B models. The task-specific checkpoints are fine-tuned for coding, math, and multilingual tasks;
normalized scores and their average are reported.}
\label{tab:llm_llama31_8b}
\centering
\small
\setlength{\tabcolsep}{2.5pt}
\resizebox{\linewidth}{!}{%
\begin{tabular}{lc cccccc c}
\toprule
\textbf{Method} & with \textbf{CUSD} & \textbf{Humaneval+} & \textbf{MBPP+} & \textbf{GSM8k} & \textbf{M-MMLU} & \textbf{M-ARC} & \textbf{M-HellaSwag} & \textbf{Avg.} \\
\midrule
Fine-tuned & --- & \resscore{100.0} & \resscore{100.0} & \resscore{100.0} & \resscore{100.0} & \resscore{100.0} & \resscore{100.0} & \resscore{100.0} \\
\midrule
\multirow{2}{*}{Task Arithmetic}
 & \resx & \resscore{82.5} & \resscore{100.0} & \resscore{93.7} & \resscore{95.5} & \resscore{89.7} & \resscore{100.8} & \resscore{93.7} \\
 & \rescheck & \resup{82.6}{0.1} & \resup{102.4}{2.4} & \resdown{91.3}{2.4} & \resup{96.1}{0.6} & \resup{90.9}{1.2} & \resup{101.3}{0.5} & \resup{\textbf{94.1}}{0.4} \\
\midrule
\multirow{2}{*}{TIES-Merging}
 & \resx & \resscore{82.6} & \resscore{100.8} & \resscore{94.2} & \resscore{95.2} & \resscore{88.8} & \resscore{100.4} & \resscore{93.7} \\
 & \rescheck & \resup{83.0}{0.4} & \resup{104.0}{3.2} & \resdown{88.2}{6.0} & \resup{97.1}{1.9} & \resup{91.2}{2.4} & \resup{101.2}{0.8} & \resup{\textbf{94.1}}{0.4} \\
\midrule
\multirow{2}{*}{DARE}
 & \resx & \resscore{84.6} & \resscore{101.2} & \resscore{92.7} & \resscore{95.1} & \resscore{90.3} & \resscore{100.7} & \resscore{\textbf{94.1}} \\
 & \rescheck & \resdown{83.2}{1.4} & \resdown{100.4}{0.8} & \resdown{91.2}{1.5} & \resup{96.0}{0.9} & \resup{91.0}{0.7} & \resup{101.2}{0.5} & \resdown{93.8}{0.3} \\
\midrule
\multirow{2}{*}{Consensus}
 & \resx & \resscore{83.8} & \resscore{99.2} & \resscore{89.5} & \resscore{97.1} & \resscore{92.3} & \resscore{101.1} & \resscore{93.8} \\
 & \rescheck & \resdown{83.4}{0.4} & \resup{100.8}{1.6} & \resup{91.9}{2.4} & \resdown{96.2}{0.9} & \resdown{90.4}{1.9} & \resdown{100.9}{0.2} & \resup{\textbf{93.9}}{0.1} \\
\bottomrule
\end{tabular}
}
\end{table}

\begin{table}[H]
\caption{Performance of merged Llama-3.2-3B-Instruct models. The task-specific checkpoints are fine-tuned for coding, math, and multilingual tasks; normalized scores and their average are reported.}
\label{tab:llm_llama32_3b_instruct}
\centering
\small
\setlength{\tabcolsep}{2.5pt}
\resizebox{\linewidth}{!}{%
\begin{tabular}{lc cccccc c}
\toprule
\textbf{Method} & with \textbf{CUSD} & \textbf{Humaneval+} & \textbf{MBPP+} & \textbf{GSM8k} & \textbf{M-MMLU} & \textbf{M-ARC} & \textbf{M-HellaSwag} & \textbf{Avg.} \\
\midrule
Fine-tuned & --- & \resscore{100.0} & \resscore{100.0} & \resscore{100.0} & \resscore{100.0} & \resscore{100.0} & \resscore{100.0} & \resscore{100.0} \\
\midrule
\multirow{2}{*}{Task Arithmetic}
 & \resx & \resscore{55.5} & \resscore{46.5} & \resscore{64.2} & \resscore{85.5} & \resscore{96.2} & \resscore{93.3} & \resscore{73.5} \\
 & \rescheck & \resup{61.3}{5.8} & \resup{54.0}{7.5} & \resdown{62.4}{1.8} & \resup{86.4}{0.9} & \resup{97.2}{1.0} & \resup{93.6}{0.3} & \resup{\textbf{75.8}}{2.3} \\
\midrule
\multirow{2}{*}{TIES-Merging}
 & \resx & \resscore{53.8} & \resscore{44.3} & \resscore{63.9} & \resscore{85.1} & \resscore{95.9} & \resscore{93.4} & \resscore{72.7} \\
 & \rescheck & \resup{58.0}{4.2} & \resup{57.5}{13.2} & \resdown{61.1}{2.8} & \resup{86.4}{1.3} & \resup{96.4}{0.5} & \resup{93.6}{0.2} & \resup{\textbf{75.5}}{2.8} \\
\midrule
\multirow{2}{*}{DARE}
 & \resx & \resscore{51.7} & \resscore{50.0} & \resscore{64.6} & \resscore{85.0} & \resscore{95.9} & \resscore{93.3} & \resscore{73.4} \\
 & \rescheck & \resup{57.6}{5.9} & \resup{56.6}{6.6} & \resdown{62.4}{2.2} & \resup{85.9}{0.9} & \resup{96.3}{0.4} & \resup{93.4}{0.1} & \resup{\textbf{75.4}}{2.0} \\
\midrule
\multirow{2}{*}{Consensus}
 & \resx & \resscore{57.8} & \resscore{59.3} & \resscore{63.8} & \resscore{86.3} & \resscore{96.5} & \resscore{93.8} & \resscore{76.2} \\
 & \rescheck & \resup{60.2}{2.4} & \resup{60.2}{0.9} & \resdown{62.5}{1.3} & \resup{86.5}{0.2} & \resdown{96.4}{0.1} & \resdown{93.7}{0.1} & \resup{\textbf{76.6}}{0.4} \\
\bottomrule
\end{tabular}
}
\end{table}

\begin{table}[H]
\caption{Performance of merged Llama-3.1-8B-Instruct models. The task-specific checkpoints are fine-tuned for coding, math, and multilingual tasks; normalized scores and their average are reported.}
\label{tab:llm_llama31_8b_instruct}
\centering
\small
\setlength{\tabcolsep}{2.5pt}
\resizebox{\linewidth}{!}{%
\begin{tabular}{lc cccccc c}
\toprule
\textbf{Method} & with \textbf{CUSD} & \textbf{Humaneval+} & \textbf{MBPP+} & \textbf{GSM8k} & \textbf{M-MMLU} & \textbf{M-ARC} & \textbf{M-HellaSwag} & \textbf{Avg.} \\
\midrule
Fine-tuned & --- & \resscore{100.0} & \resscore{100.0} & \resscore{100.0} & \resscore{100.0} & \resscore{100.0} & \resscore{100.0} & \resscore{100.0} \\
\midrule
\multirow{2}{*}{Task Arithmetic}
 & \resx & \resscore{55.5} & \resscore{87.9} & \resscore{87.6} & \resscore{84.2} & \resscore{87.3} & \resscore{92.4} & \resscore{82.5} \\
 & \rescheck & \resup{70.7}{15.2} & \resup{91.4}{3.5} & \resdown{83.8}{3.8} & \resup{88.3}{4.1} & \resup{90.7}{3.4} & \resup{94.0}{1.6} & \resup{\textbf{86.5}}{4.0} \\
\midrule
\multirow{2}{*}{TIES-Merging}
 & \resx & \resscore{56.9} & \resscore{87.9} & \resscore{87.8} & \resscore{84.5} & \resscore{86.7} & \resscore{92.1} & \resscore{82.6} \\
 & \rescheck & \resup{68.8}{11.9} & \resup{92.6}{4.7} & \resdown{85.2}{2.6} & \resup{87.3}{2.8} & \resup{89.6}{2.9} & \resup{93.8}{1.7} & \resup{\textbf{86.2}}{3.6} \\
\midrule
\multirow{2}{*}{DARE}
 & \resx & \resscore{59.7} & \resscore{87.5} & \resscore{87.9} & \resscore{83.8} & \resscore{87.3} & \resscore{92.3} & \resscore{83.1} \\
 & \rescheck & \resup{71.7}{12.0} & \resup{91.8}{4.3} & \resdown{84.9}{3.0} & \resup{86.8}{3.0} & \resup{89.5}{2.2} & \resup{93.7}{1.4} & \resup{\textbf{86.4}}{3.3} \\
\midrule
\multirow{2}{*}{Consensus}
 & \resx & \resscore{68.6} & \resscore{93.4} & \resscore{85.5} & \resscore{86.6} & \resscore{89.2} & \resscore{93.7} & \resscore{86.2} \\
 & \rescheck & \resup{73.7}{5.1} & \resup{94.9}{1.5} & \resdown{82.7}{2.8} & \resup{88.7}{2.1} & \resup{91.2}{2.0} & \resup{94.0}{0.3} & \resup{\textbf{87.5}}{1.3} \\
\bottomrule
\end{tabular}
}
\end{table}

\begin{table}[H]
\caption{Performance of merged Gemma-2-2b models. The task-specific checkpoints are fine-tuned for coding, math, and multilingual tasks; normalized scores and their average are reported.}
\label{tab:llm_gemma2_2b}
\centering
\small
\setlength{\tabcolsep}{2.5pt}
\resizebox{\linewidth}{!}{%
\begin{tabular}{lc cccccc c}
\toprule
\textbf{Method} & with \textbf{CUSD} & \textbf{Humaneval+} & \textbf{MBPP+} & \textbf{GSM8k} & \textbf{M-MMLU} & \textbf{M-ARC} & \textbf{M-HellaSwag} & \textbf{Avg.} \\
\midrule
Fine-tuned & --- & \resscore{100.0} & \resscore{100.0} & \resscore{100.0} & \resscore{100.0} & \resscore{100.0} & \resscore{100.0} & \resscore{100.0} \\
\midrule
\multirow{2}{*}{Task Arithmetic}
 & \resx & \resscore{80.6} & \resscore{94.9} & \resscore{75.0} & \resscore{106.8} & \resscore{104.0} & \resscore{104.0} & \resscore{94.2} \\
 & \rescheck & \resup{83.7}{3.1} & \resup{98.0}{3.1} & \resup{76.4}{1.4} & \resup{107.4}{0.6} & \resup{104.5}{0.5} & \resup{104.1}{0.1} & \resup{\textbf{95.7}}{1.5} \\
\midrule
\multirow{2}{*}{TIES-Merging}
 & \resx & \resscore{80.4} & \resscore{90.3} & \resscore{79.5} & \resscore{107.1} & \resscore{103.3} & \resscore{103.8} & \resscore{94.1} \\
 & \rescheck & \resup{90.8}{10.4} & \resup{94.4}{4.1} & \resup{83.2}{3.7} & \resup{107.6}{0.5} & \resup{104.0}{0.7} & \resup{103.9}{0.1} & \resup{\textbf{97.3}}{3.2} \\
\midrule
\multirow{2}{*}{DARE}
 & \resx & \resscore{80.6} & \resscore{91.8} & \resscore{76.7} & \resscore{107.6} & \resscore{104.0} & \resscore{104.2} & \resscore{94.1} \\
 & \rescheck & \resup{87.3}{6.7} & \resup{94.9}{3.1} & \resup{78.6}{1.9} & \resup{108.1}{0.5} & \resup{104.1}{0.1} & \resdown{104.1}{0.1} & \resup{\textbf{96.2}}{2.1} \\
\midrule
\multirow{2}{*}{Consensus}
 & \resx & \resscore{60.5} & \resscore{94.9} & \resscore{68.8} & \resscore{103.2} & \resscore{102.7} & \resscore{102.8} & \resscore{88.8} \\
 & \rescheck & \resup{77.5}{17.0} & \resdown{90.3}{4.6} & \resup{75.8}{7.0} & \resup{107.2}{4.0} & \resup{104.3}{1.6} & \resup{103.7}{0.9} & \resup{\textbf{93.1}}{4.3} \\
\bottomrule
\end{tabular}
}
\end{table}

\begin{table}[H]
\caption{Performance of merged Gemma-2-9b models. The task-specific checkpoints are fine-tuned for coding, math, and multilingual tasks;normalized scores and their average are reported.}
\label{tab:llm_gemma2_9b}
\centering
\small
\setlength{\tabcolsep}{2.5pt}
\resizebox{\linewidth}{!}{%
\begin{tabular}{lc cccccc c}
\toprule
\textbf{Method} & with \textbf{CUSD} & \textbf{Humaneval+} & \textbf{MBPP+} & \textbf{GSM8k} & \textbf{M-MMLU} & \textbf{M-ARC} & \textbf{M-HellaSwag} & \textbf{Avg.} \\
\midrule
Fine-tuned & --- & \resscore{100.0} & \resscore{100.0} & \resscore{100.0} & \resscore{100.0} & \resscore{100.0} & \resscore{100.0} & \resscore{100.0} \\
\midrule
\multirow{2}{*}{Task Arithmetic}
 & \resx & \resscore{73.5} & \resscore{102.8} & \resscore{105.0} & \resscore{96.3} & \resscore{83.4} & \resscore{90.7} & \resscore{92.0} \\
 & \rescheck & \resdown{72.4}{1.1} & \resup{106.8}{4.0} & \resdown{101.3}{3.7} & \resup{99.5}{3.2} & \resup{90.7}{7.3} & \resup{98.1}{7.4} & \resup{\textbf{94.8}}{2.8} \\
\midrule
\multirow{2}{*}{TIES-Merging}
 & \resx & \resscore{71.5} & \resscore{102.0} & \resscore{105.0} & \resscore{96.2} & \resscore{81.8} & \resscore{90.3} & \resscore{91.1} \\
 & \rescheck & \ressame{71.5}{0.0} & \resup{109.2}{7.2} & \resdown{99.6}{5.4} & \resup{100.3}{4.1} & \resup{91.5}{9.7} & \resup{99.6}{9.3} & \resup{\textbf{95.3}}{4.2} \\
\midrule
\multirow{2}{*}{DARE}
 & \resx & \resscore{72.8} & \resscore{99.6} & \resscore{105.5} & \resscore{96.0} & \resscore{82.9} & \resscore{90.3} & \resscore{91.2} \\
 & \rescheck & \resdown{72.1}{0.7} & \resup{106.8}{7.2} & \resdown{101.7}{3.8} & \resup{99.5}{3.5} & \resup{90.7}{7.8} & \resup{98.2}{7.9} & \resup{\textbf{94.8}}{3.6} \\
\midrule
\multirow{2}{*}{Consensus}
 & \resx & \resscore{80.7} & \resscore{105.2} & \resscore{100.2} & \resscore{99.3} & \resscore{88.6} & \resscore{93.7} & \resscore{94.6} \\
 & \rescheck & \resdown{74.4}{6.3} & \resup{107.6}{2.4} & \resup{100.6}{0.4} & \resup{100.3}{1.0} & \resup{91.7}{3.1} & \resup{98.5}{4.8} & \resup{\textbf{95.5}}{0.9} \\
\bottomrule
\end{tabular}
}
\end{table}

\subsection{Dispersion Scoring Analysis}

\begin{figure}[t]
    \centering
    \includegraphics[width=\textwidth]{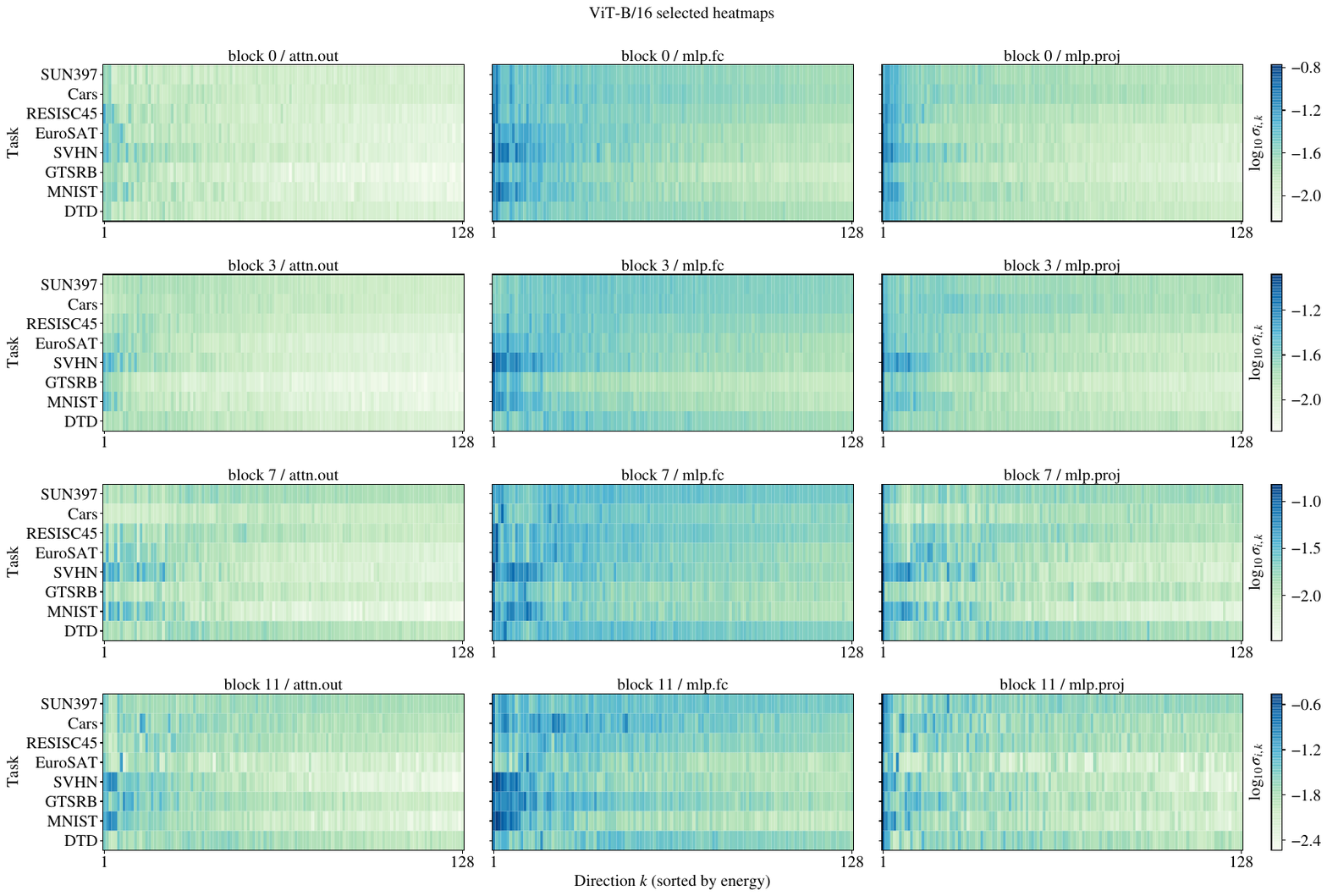}

    \caption{\textbf{ViT-B/16 directional singular-value heatmaps.} Each panel shows $\log_{10}\sigma_{i,k}$ for one layer component, with directions sorted by energy.}

    \label{fig:sigma_heatmap_vit_b16_appendix}
  \end{figure}

\begin{figure}[t]
    \centering
    \includegraphics[width=\textwidth]{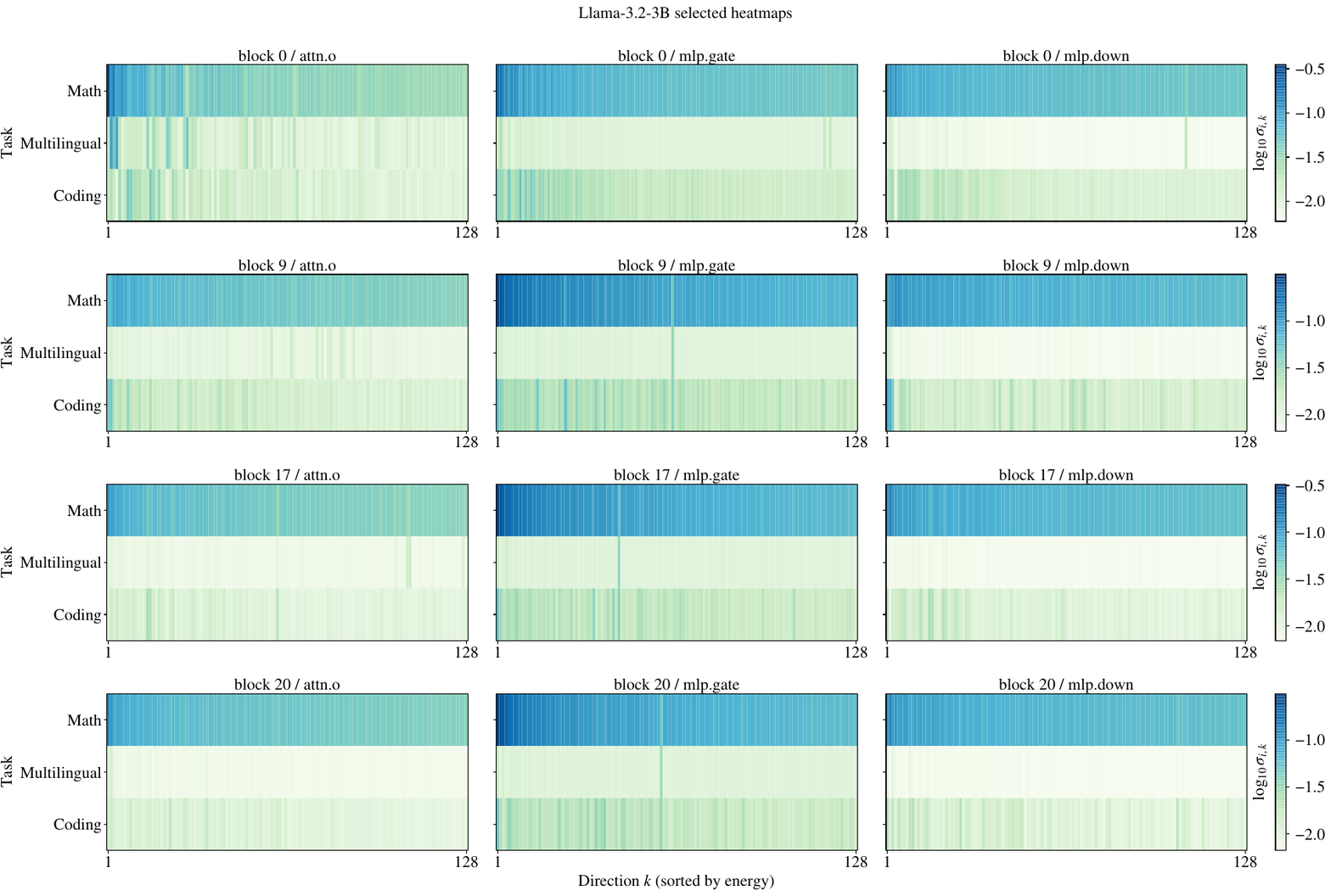}

    \caption{\textbf{Llama-3.2-3B directional singular-value heatmaps.} Each panel shows $\log_{10}\sigma_{i,k}$ for one layer component, with directions sorted by energy.}
    
    \label{fig:sigma_heatmap_llama_3b_appendix}
  \end{figure}

Figure~\ref{fig:sigma_heatmap_vit_b16_appendix} and Figure~\ref{fig:sigma_heatmap_llama_3b_appendix} provide additional layer heatmaps for the representative ViT and Llama settings used in the dispersion-scoring analysis.

\subsection{Additional Hyperparameter Sensitivity Analysis}
\label{app:beta_consistency}

\begin{figure}[H]
\centering
\includegraphics[width=0.95\textwidth]{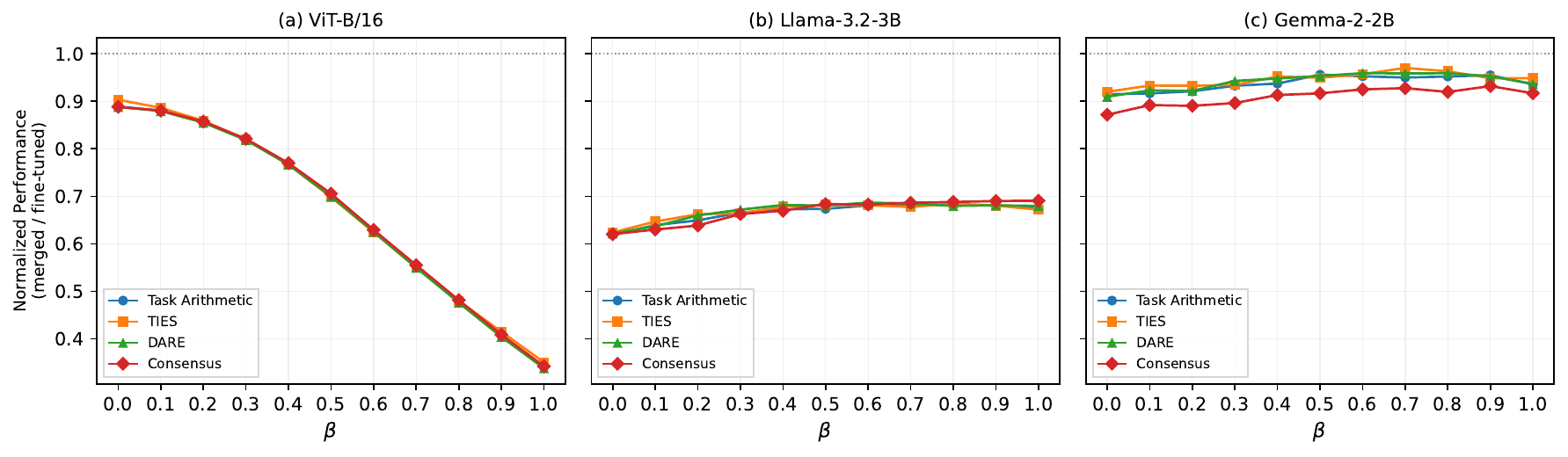}
\caption{Per-method sensitivity to $\beta$ across downstream arithmetic-based merging operators.}
\label{fig:beta_per_method_consistency}
\end{figure}

Figure~\ref{fig:beta_per_method_consistency} reports $\beta$ sensitivity separately for each downstream merging method. This complements the averaged sweep in Figure~\ref{fig:beta_sweep} and checks whether the observed trend is consistent across Task Arithmetic, TIES-Merging, DARE, and Consensus Merging.

\begin{table}[t]
\centering
\caption{Joint sensitivity to the Task Arithmetic scale $\lambda$
and unique-residual coefficient $\beta$ on ViT-B/32.
We report average accuracy (\%) from a post-hoc sweep on the
test sets. The final column reports plain TA at the same $\lambda$;
bold entries outperform this baseline.
$^{\dagger}$ denotes the hyperparameter configuration used in
the original submission.}
\label{tab:ta_joint_sweep}
\small
\setlength{\tabcolsep}{2.5pt}
\resizebox{\linewidth}{!}{%
\begin{tabular}{c*{11}{r}r}
\toprule
$\lambda \backslash \beta$
& 0.0 & 0.1 & 0.2 & 0.3 & 0.4 & 0.5
& 0.6 & 0.7 & 0.8 & 0.9 & 1.0 & Plain TA \\
\midrule
0.1
& 57.43 & 61.74 & \textbf{63.33} & \textbf{62.22}
& 59.15 & 53.95 & 47.22 & 39.87 & 33.62 & 28.21 & 23.91
& 61.74 \\
0.2
& 66.62 & 69.07 & 69.36 & 67.35
& 62.75 & 56.56 & 49.11 & 41.37 & 34.97 & 29.31 & 24.91
& 69.36 \\
0.3
& \textbf{72.49} & \textbf{73.87} & \textbf{72.94} & 70.05
& 64.59 & 57.76 & 49.88 & 42.29 & 35.91 & 30.36 & 25.85
& 70.05 \\
0.4
& \textbf{76.48}$^{\dagger}$
& \textbf{76.50} & \textbf{74.65} & \textbf{70.48}
& 64.65 & 57.43 & 49.84 & 42.52 & 36.31 & 31.18 & 26.94
& 64.65$^{\dagger}$ \\
0.5
& \textbf{77.64} & \textbf{77.02} & \textbf{74.15}
& \textbf{69.70} & \textbf{63.45} & 56.27
& 49.03 & 42.36 & 36.58 & 31.94 & 27.80
& 56.27 \\
0.6
& \textbf{77.33} & \textbf{75.82} & \textbf{72.47}
& \textbf{67.53} & \textbf{61.25} & \textbf{54.54}
& 47.97 & 41.74 & 36.49 & 32.35 & 28.43
& 47.97 \\
0.7
& \textbf{75.53} & \textbf{73.54} & \textbf{69.82}
& \textbf{64.63} & \textbf{58.90} & \textbf{52.56}
& \textbf{46.45} & 40.91 & 36.36 & 32.47 & 28.86
& 40.91 \\
0.8
& \textbf{72.48} & \textbf{70.22} & \textbf{66.47}
& \textbf{61.61} & \textbf{56.11} & \textbf{50.44}
& \textbf{45.04} & \textbf{40.05} & 36.13 & 32.33 & 28.99
& 36.13 \\
0.9
& \textbf{68.78} & \textbf{66.37} & \textbf{62.71}
& \textbf{58.20} & \textbf{53.25} & \textbf{48.21}
& \textbf{43.42} & \textbf{39.15} & \textbf{35.57}
& 32.20 & 29.06
& 32.20 \\
1.0
& \textbf{64.46} & \textbf{62.11} & \textbf{58.83}
& \textbf{54.94} & \textbf{50.45} & \textbf{46.04}
& \textbf{41.90} & \textbf{38.28} & \textbf{34.92}
& \textbf{31.87} & 28.98
& 28.98 \\
\bottomrule
\end{tabular}%
}
\end{table}

\subsection{Additional Per-Block Interference Analysis}
\label{app:interference_analysis}

\begin{figure}[H]
\centering
\begin{subfigure}[t]{0.32\linewidth}
  \centering
  \includegraphics[width=\linewidth]{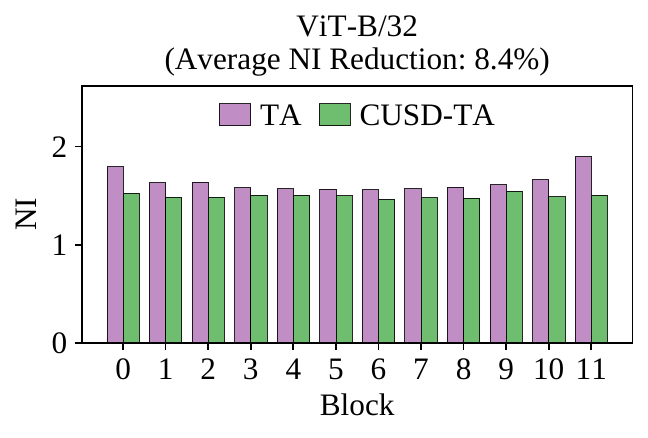}
  \caption{ViT-B/32.}
\end{subfigure}
\hfill
\begin{subfigure}[t]{0.32\linewidth}
  \centering
  \includegraphics[width=\linewidth]{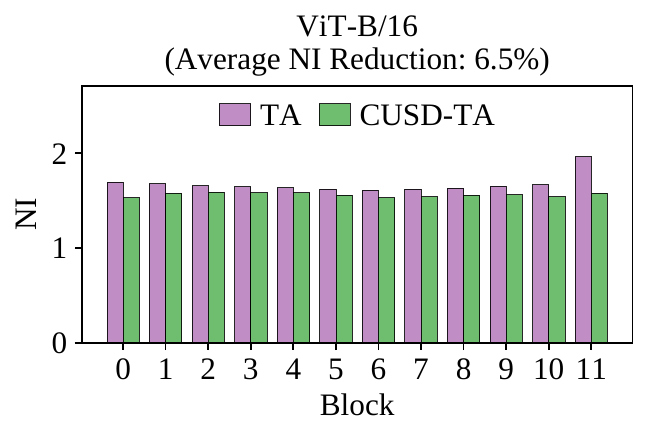}
  \caption{ViT-B/16.}
\end{subfigure}
\hfill
\begin{subfigure}[t]{0.32\linewidth}
  \centering
  \includegraphics[width=\linewidth]{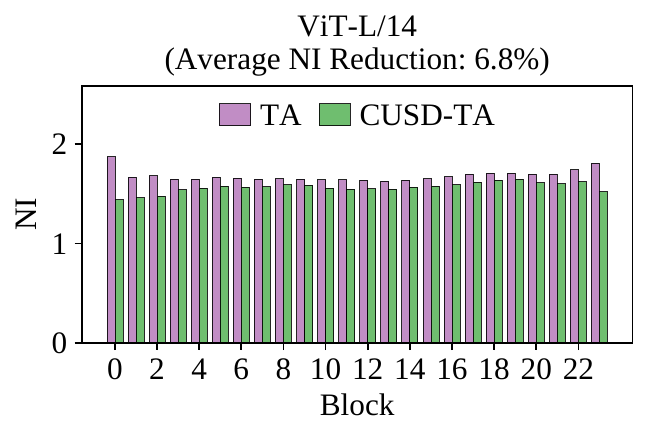}
  \caption{ViT-L/14.}
\end{subfigure}
\caption{Per-block interference analysis for Task Arithmetic on ViT models.}
\label{fig:interference_ta_vit}
\end{figure}

\begin{figure}[H]
\centering
\begin{subfigure}[t]{0.32\linewidth}
  \centering
  \includegraphics[width=\linewidth]{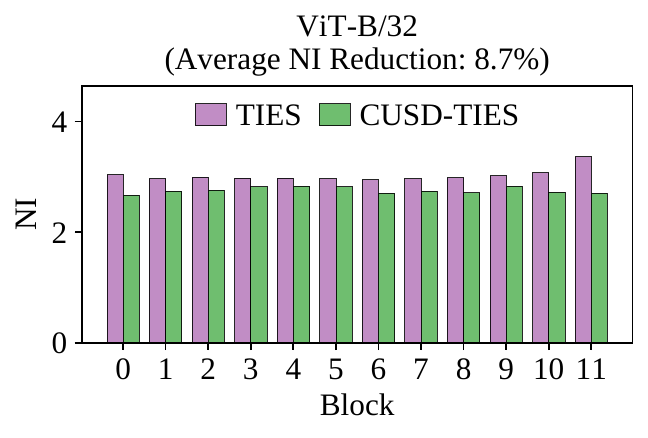}
  \caption{ViT-B/32.}
\end{subfigure}
\hfill
\begin{subfigure}[t]{0.32\linewidth}
  \centering
  \includegraphics[width=\linewidth]{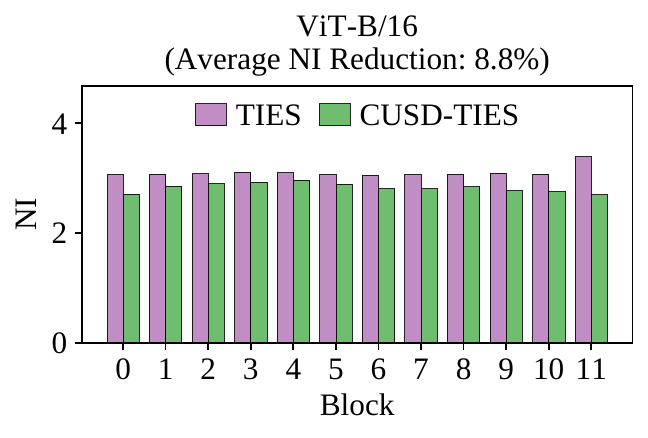}
  \caption{ViT-B/16.}
\end{subfigure}
\hfill
\begin{subfigure}[t]{0.32\linewidth}
  \centering
  \includegraphics[width=\linewidth]{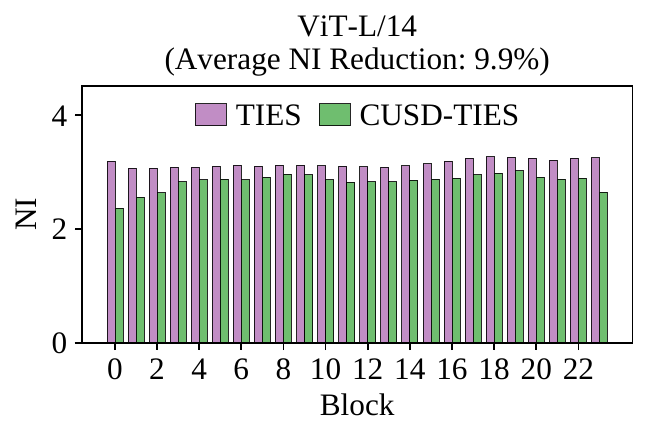}
  \caption{ViT-L/14.}
\end{subfigure}
\caption{Per-block interference analysis for TIES-Merging on ViT models.}
\label{fig:interference_ties_vit}
\end{figure}

\begin{figure}[H]
\centering
\begin{subfigure}[t]{0.32\linewidth}
  \centering
  \includegraphics[width=\linewidth]{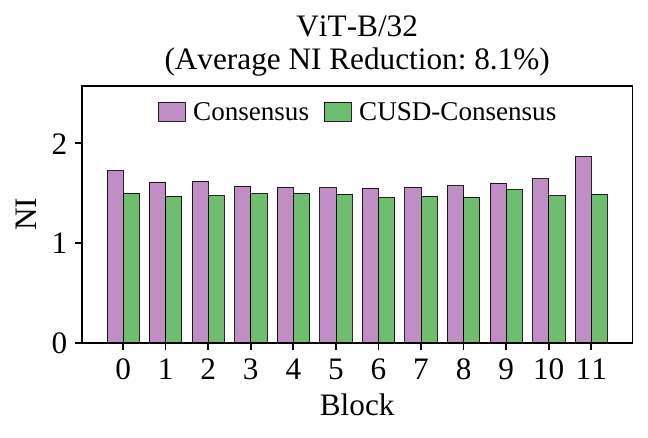}
  \caption{ViT-B/32.}
\end{subfigure}
\hfill
\begin{subfigure}[t]{0.32\linewidth}
  \centering
  \includegraphics[width=\linewidth]{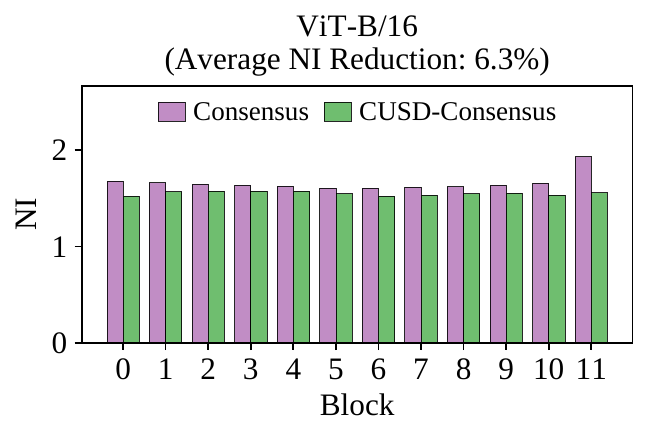}
  \caption{ViT-B/16.}
\end{subfigure}
\hfill
\begin{subfigure}[t]{0.32\linewidth}
  \centering
  \includegraphics[width=\linewidth]{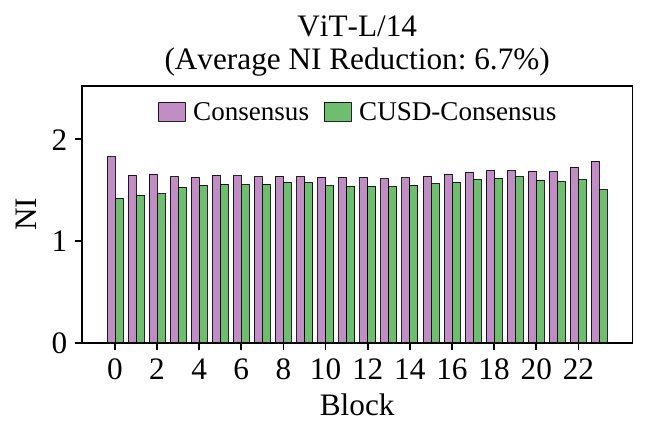}
  \caption{ViT-L/14.}
\end{subfigure}
\caption{Per-block interference analysis for Consensus Merging on ViT models.}
\label{fig:interference_consensus_vit}
\end{figure}

For a merged task vector $\tau_m^{(\ell)}$ at layer $\ell$ and task-specific task vectors $\{\tau_i^{(\ell)}\}_{i=1}^K$, we measure a normalized interference proxy (NI) as
\begin{equation}
  \mathrm{NI}
  =
  \frac{1}{K}
  \sum_{i=1}^{K}
  \frac{
    \left\|\tau_m - \tau_i\right\|_F^2
  }{
    \left\|\tau_i\right\|_F^2
  }.
  \label{eq:normalized_interference}
\end{equation}
For the per-block plots, we aggregate the decomposed matrix-structured layers in each block. We use this normalized distance as a parameter-space proxy for interference: a larger value means that the merged update deviates more from the task-specific updates. Figure~\ref{fig:interference_ta_vit}, Figure~\ref{fig:interference_ties_vit}, and Figure~\ref{fig:interference_consensus_vit} provide the per-block interference analysis for the remaining merging methods on ViT models.

\subsection{Limitations}
\label{app:limitation}

We acknowledge the potential limitations of our work. Performance gains vary across methods and models, and
SVD-based baselines remain stronger in some evaluated settings.
CUSD uses only model weights during decomposition and merging,
but selecting $\beta$ can require validation data. CUSD is evaluated in the standard homogeneous-merging setting, where all task-specific checkpoints are fine-tuned from the same base architecture. It does not address heterogeneous merging across different architectures. Our evaluation also focuses on dense models at the ViT, RoBERTa, Llama, and Gemma scales considered in this work; substantially larger checkpoints, such as 70B-class LLMs, and Mixture-of-Experts architectures are left for future work. Future work could also explore multiobjective optimization
for balancing common aggregation and unique
retention~\citep{wang2023extended,zou2023evolutionary}.
We further plan to investigate merging compatible models
for RNA design~\citep{hu2026rider,hu2026cocorna,ma2026reflex,hu2026tango}.

\subsection{Broader Impacts}
\label{app:broader_impacts}

  \textbf{Positive Societal Impacts}\quad By improving data-free arithmetic model merging, CUSD has the potential for several positive societal
   impacts. Because CUSD uses only model weights during decomposition
and merging, it can support privacy-sensitive and federated
settings where checkpoints are produced by different owners
and the underlying data cannot be shared, e.g., when fine-tuned models from hospitals, financial institutions, or independent research groups must be combined into a
   single multi-task model without disclosing proprietary or personal data. By more reliably merging task-specific checkpoints into one
  multi-task model, CUSD reduces the need to deploy and serve a separate fine-tuned model per task, which lowers storage, memory, inference
  compute, and energy consumption, and is particularly beneficial in resource-constrained environments such as edge devices and small
  organizations. Because CUSD operates as a pre-merging module that augments existing arithmetic operators (Task Arithmetic, TIES, DARE,
  Consensus) rather than replacing them, practitioners can adopt it without retraining or redesigning their pipelines, which broadens access to
   high-quality multi-task model assembly across the ViT, RoBERTa, Llama, and Gemma families considered in this work and similar architectures.
   Together, these properties can accelerate research and deployment of capable multi-task models in domains such as scientific discovery,
  medical decision support, multilingual NLP, and educational tools, and can make advanced AI capabilities available to a broader range of
  users and institutions.

  \textbf{Negative Societal Impacts}\quad Despite these benefits, the advances presented also carry potential negative societal impacts.
  Improved data-free merging makes it easier to combine the capabilities of independently distributed checkpoints, which could be leveraged for
   malicious purposes, such as merging a safety-aligned model with a checkpoint fine-tuned for harmful or unsafe behavior in order to
  circumvent alignment, or assembling specialized capabilities for generating disinformation, sophisticated phishing content, or deceptive
  media at scale. Because CUSD merges models without inspecting their training data, biases, unsafe behaviors, or memorized content present in
  any constituent checkpoint can be inherited and aggregated in the merged model, potentially leading to unfair or discriminatory outcomes when
   these models are deployed in decision-making pipelines. The reduced cost of obtaining strong multi-task models from publicly available
  checkpoints might also lower the barrier to repurposing released models in ways that their original developers did not intend, and may
  contribute to the concentration of capabilities in the hands of actors with the largest pools of high-quality fine-tuned weights. Mitigation
  strategies—such as releasing merging tools alongside responsible-use guidelines, evaluating merged models for safety, fairness, and bias
  before deployment, developing detectors for content produced by merged models, and applying safety filters or alignment procedures after
  merging—are important to address these risks.


\clearpage
\input{checklist.tex}

\end{document}

%% file: sections/introduction.tex

\section{Introduction}
\label{sec:introduction}

Model merging aims to combine $K$ task-specific checkpoints, fine-tuned from a common base model, into a single multi-task model without access to the original task data or joint retraining~\citep{yang2026model, song2026model, yaoactivation, wang2025model, nobariactivation,hameed2026model, xu2026ace, sun2025task}. This setting is increasingly relevant when training data cannot be shared, checkpoints are produced by different owners, or serving one model per task is too expensive~\citep{roth2024practitioner, yadav2024matters}. Arithmetic-based merging methods are a central paradigm in this setting because they require only model weights. They represent each task by a task vector $\tau_i=\theta_i-\theta_{\mathrm{base}}$~\citep{ilharco2022editing} and construct a merged update using non-iterative operations such as addition, sign selection, sparsification, or rescaling~\citep{ilharco2022editing, wortsman2022model, yadav2023ties, yu2024language, wang2024localizing}. Despite substantial progress~\citep{marczak2025no, wang2025lines}, arithmetic merging often remains below the corresponding task-specific models, indicating that residual inter-task interference is not fully resolved by current rules.

We identify one source of this interference: current arithmetic methods are largely coordinate-wise and uniform. They decide how to add, trim, drop, or rescale scalar entries of task vectors, but they do not first determine whether those entries belong to update directions that are consistently shared across tasks or dominated by only a few tasks. This distinction matters because parameter updates are not merely collections of independent coordinates. In matrix-structured layers, a direction represents a coordinated update pattern across many parameter entries. A direction used comparably by all tasks is a candidate for aggregation, whereas a direction dominated by one task may need to be preserved, damped, or separated from the shared merge. Applying the same arithmetic rule to both types of directions can either suppress task-dominated signal or inject task-specific interference into the common update.

\begin{figure}[t]
\centering
\includegraphics[width=\textwidth]{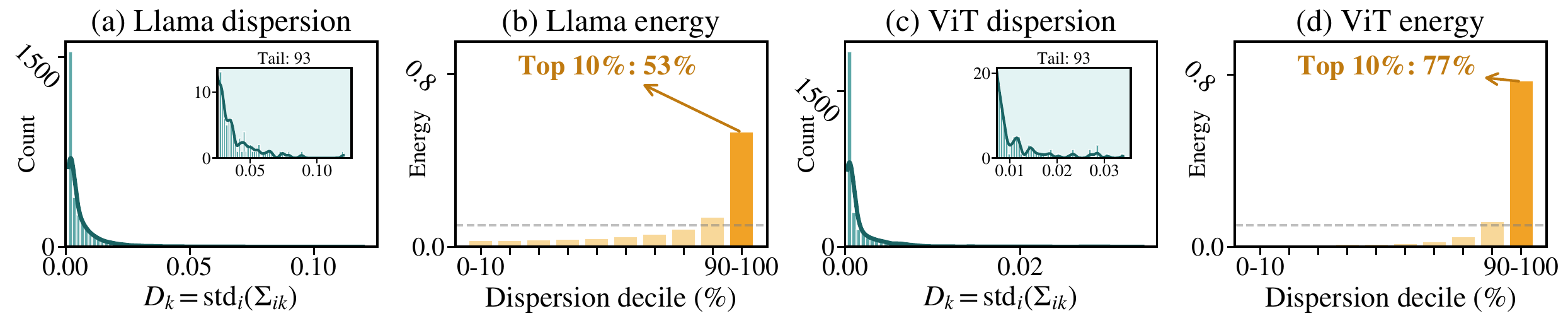}
\caption{\textbf{Direction-level heterogeneity in task vectors.}
Motivating examples from representative matrix layers of Llama-3.2-3B (\texttt{mlp.gate\_proj}) and ViT-B/16 (\texttt{mlp.c\_proj}). 
(a,\,c) Distribution of cross-task dispersion scores over shared directions.
(b,\,d) Share of total directional energy in each dispersion-ranked decile.}
\label{fig:motivation}
\end{figure}

 To expose this structure, we analyze task-vector matrices in a shared directional basis. We use energy to denote the update magnitude carried by task vectors along a shared direction. For each shared direction, we measure the task-wise energy and its dispersion across tasks. Figure~\ref{fig:motivation} shows two motivating examples from Llama-3.2-3B~\citep{grattafiori2024llama} and ViT-B/16~\citep{dosovitskiy2020image}. In both cases, most directions form a low-dispersion bulk, while a small high-dispersion tail carries a disproportionate fraction of the total directional energy. In the shown layers, the highest-dispersion decile accounts for $53\%$ and $77\%$ of total energy, respectively. Thus, the high-dispersion tail is not negligible noise: it contains substantial update magnitude, but its energy is unevenly distributed across tasks. This observation motivates separating directions by their cross-task energy dispersion before applying an arithmetic merging rule. Appendix~\ref{app:directional_structure} provides additional cross-architecture visualizations showing that the same two-regime pattern appears broadly across layers, model families, and task groups.

We propose \textbf{C}ommon--\textbf{U}nique \textbf{S}ubspace \textbf{D}ecomposition (CUSD), a data-free pre-merging framework for arithmetic model merging. For each matrix-structured layer, CUSD constructs a shared orthogonal directional basis from the task-vector matrices using a regularized Higher-Order Generalized Singular Value Decomposition derived (HO-GSVD-derived) construction~\citep{ponnapalli2011higher, kempf2023higher}. In this shared basis, CUSD computes a cross-task dispersion score for each direction and partitions directions into two sets: common directions, whose energy is comparable across tasks, and unique directions, whose energy is dominated by one or a few tasks. We use ``unique'' in this operational sense: these directions are task-dominated residuals, not assumed to encode semantically exclusive features.

Given this partition, the common components $C_i$ and unique residuals $R_i$ can be obtained from $K$ task vectors. It then applies the downstream arithmetic merging operator $\mathcal{F}$ only to the $C_i$ and controls the $R_i$ with a separate scalar coefficient $\beta$ to get merged task vector $\tau_{m}$:
\[
    \tau_{m}
    =
    \mathcal{F}(C_1, \ldots, C_K)
    +
    \beta \sum\nolimits_{i=1}^K R_i , 
\]
This design makes CUSD a pre-merging module rather than a replacement for existing arithmetic operators. The same decomposition depends only on the task vectors and can therefore be computed once and reused across different arithmetic merging rules~\citep{ilharco2022editing, yu2024language, wang2024localizing}. Compared with per-task SVD methods~\citep{klema1980singular,gargiulo2025task}, CUSD avoids comparing directions chosen independently for each task; compared with post-merge subspace corrections~\citep{skorobogat2025subspace}, it separates task-dominated directions before the merging rule has mixed them; and compared with data-dependent routing or modular approaches~\citep{lu2024twin}, it produces a single static merged model using weights alone. Our main contributions are summarized as follows:
\begin{itemize}
    \item We identify and quantify a direction-level two-regime structure in task-vector matrices: a low-dispersion bulk of broadly shared directions and a high-dispersion, task-dominated tail that carries disproportionate energy.

    \item We introduce CUSD, a data-free pre-merging method that uses a HO-GSVD-derived construction to obtain a shared directional basis, separates common components from unique residuals, applies any downstream arithmetic merging operator only to the common components, and controls the unique residuals through a single coefficient $\beta$.

    \item We evaluate CUSD across $11$ base models, four model families, and four arithmetic-based merging operators, showing that CUSD consistently improves data-free arithmetic merging across the evaluated settings.
\end{itemize}

%% file: sections/related_work.tex

\section{Related Work}
\label{sec:related_work}

Current model merging research falls into two categories: data-free methods that rely only on model parameters or task vectors, and data-dependent methods that require additional data during merging.

\subsection{Data-Free Model Merging}

\paragraph{Arithmetic-based model merging.}
These methods apply uniform element-wise rules to model weights or task vectors $\{\tau_i\}$ without data or iterative optimization.
Task Arithmetic (TA)~\citep{ilharco2022editing} merges via scaled addition $\theta_{\text{merged}} = \theta_{\text{base}} + \lambda \sum\nolimits_i \tau_i$; Model Soups~\citep{wortsman2022model} averages checkpoints.
TIES-Merging~\citep{yadav2023ties} resolves sign conflicts through trimming, majority vote, and disjoint averaging.
DARE~\citep{yu2024language} randomly prunes delta parameters with rescaling.
Consensus Merging~\citep{wang2024localizing} identifies and suppresses catastrophic and selfish parameters based on cross-task agreement patterns.
Additional methods in this class include PCB-Merging~\citep{du2024parameter} and Breadcrumbs~\citep{davari2024model}.
LiNeS~\citep{wang2025lines} applies layer-increasing scalar scaling to the merged weights to prevent catastrophic forgetting as a post-merge arithmetic correction.
All these operators decide which per-parameter values to retain, drop, or rescale; none examines the directional geometry across task vectors.

\paragraph{SVD-based model merging.}
A separate line of data-free methods moves from element-wise rules to subspace operations.
TSV-Merge~\citep{gargiulo2025task} performs per-task SVD~\citep{klema1980singular}
and orthogonalizes the retained singular vectors via a
Procrustes transformation.
KnOTS~\citep{stoica2024model} uses joint SVD to align
concatenated LoRA~\citep{hu2022lora} updates before applying a merging operator.
Iso-Merging~\citep{marczak2025no} decomposes the merged TA output into common and task-specific subspaces with isotropic scaling. The common subspace is defined by the top-$k$ singular vectors
of the summed task matrices, while task-specific subspaces are
extracted from each task's residual after projecting out the
common subspace.
Subspace Boosting~\citep{skorobogat2025subspace} diagnoses rank collapse in the merged output and mitigates it by
boosting smaller singular values, using HO-GSVD~\citep{ponnapalli2011higher, kempf2023higher} as an interpretability tool. These post-merge methods tie their decomposition to a specific merge operator, so the result cannot be reused across methods. Moreover, they repair the merged output after interference has already occurred rather than preventing it.
CUSD decomposes each task vector into common and unique subspaces before any merging operator is applied. Because the decomposition depends only on $\{\tau_i\}$, it pairs with any arithmetic-based method and gives common and unique components independent hyperparameters.

\paragraph{Optimization-based model merging.}
Unlike the non-iterative operators above, methods in this class exploit intrinsic properties of task vectors to construct optimization objectives solved iteratively.
AWD~\citep{xiong2024multi} proves that inter-task interference is minimized when task vectors are mutually orthogonal, and proposes adaptive weight disentanglement to push them toward orthogonality.
DOGE~\citep{wei2025modeling} formulates merging as a constrained optimization problem, interprets task vectors as accumulated gradients, and solves it via adaptive projected gradient descent.
WUDI-Merging~\citep{cheng2025whoever} shows that a linear layer's task vectors span an approximate linear subspace of the corresponding inputs, and uses this property to construct a data-free loss that minimizes the interference upper bound.
DOP~\citep{yang2025continual} combines data-space approximation with orthogonal projection and designs a data-free loss balancing stability and plasticity for continual merging.
Each method optimizes toward a specific structural prior, such as orthogonality (AWD), gradient alignment (DOGE), or subspace preservation (WUDI, DOP), and its effectiveness depends on how well that prior matches the actual task-vector geometry.

\subsection{Data-Dependent Model Merging}

A second category of methods uses data-derived signals, including Fisher information~\citep{rissanen2002fisher}, activation statistics, or unlabeled test samples, to guide the merging process.
Fisher Merging~\citep{matena2022merging} weights parameters by diagonal Fisher information estimated from labeled training data.
RegMean~\citep{jin2022dataless} minimizes a linear regression reconstruction loss using activation inner-product matrices.
ZipIt~\citep{stoica2023zipit} computes activation statistics on unlabeled data to align and fuse redundant features across networks.
AdaMerging~\citep{yang2023adamerging} learns task-wise or layer-wise merging coefficients via entropy minimization on unlabeled test samples.
Representation Surgery~\citep{yang2024representation} learns a linear transformation to repair merged representations using unsupervised data.
Twin-Merging~\citep{lu2024twin} decomposes each task vector via per-task SVD into shared and exclusive components, then trains a test-time router on task data to dynamically weight the exclusive components at inference.
These methods require Fisher statistics, activations, or unlabeled samples beyond model weights alone and address a different setting from the data-free problem considered here.

%% file: sections/method.tex
\section{Method}
\label{sec:method}

\begin{figure}[h]
\centering
\includegraphics[width=\textwidth]{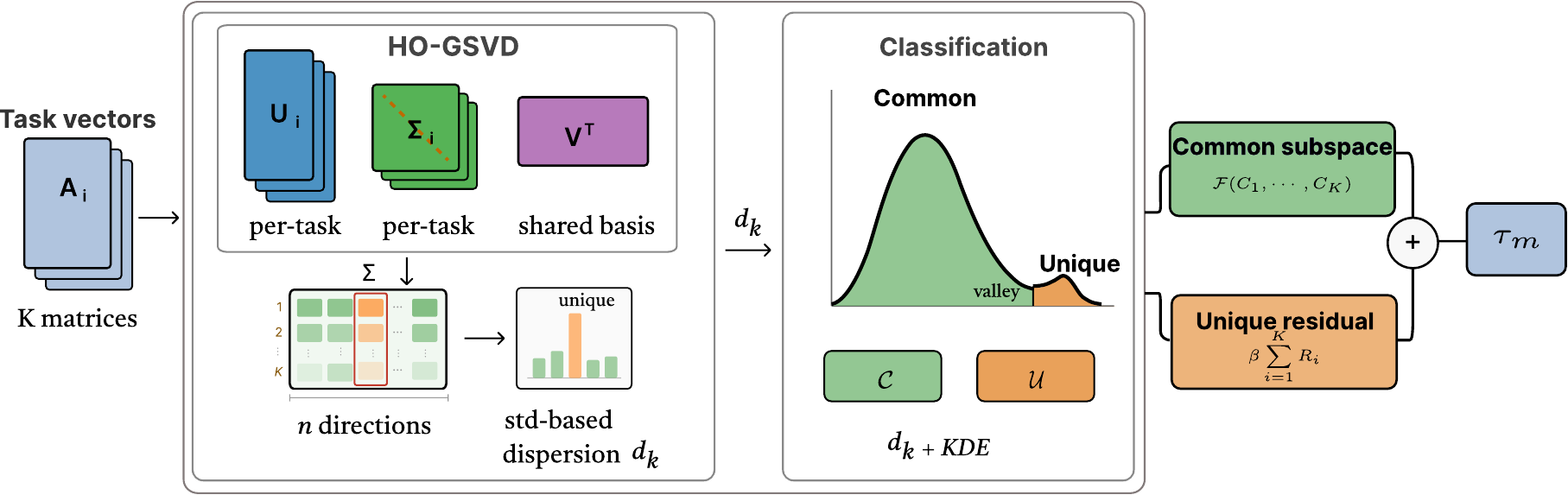}
\caption{CUSD pipeline overview. Task vectors $A_i$ are projected into an HO-GSVD-derived shared right basis $V$ and directional energies $\Sigma$. A std-based dispersion score with a KDE-based threshold partitions directions into common ($\mathcal{C}$) and unique ($\mathcal{U}$). CUSD merges common components with $\mathcal{F}$ and retains unique residuals through $\beta \sum_{i=1}^K R_i$.}
\label{fig:overview}
\end{figure}

\subsection{Preliminaries}
\label{sec:task_vectors}

Given a pre-trained base model~$\theta_{\text{base}}$ and $K$ task-specific models $\theta_1, \ldots, \theta_K$ fine-tuned on tasks $1, \ldots, K$, the task vector for task~$i$ is $\tau_i = \theta_i - \theta_{\text{base}}$~\citep{ilharco2022editing}.
Arithmetic-based merging can be written as
\begin{equation}
  \tau_m = \mathcal{F}(\tau_1, \ldots, \tau_K),
  \qquad
  \theta_{\text{merged}} = \theta_{\text{base}} + \tau_m ,
\label{eq:merge}
\end{equation}
where $\tau_m$ denotes the merged task vector, $\mathcal{F}$ denotes a downstream merging operator and $\theta_{\text{merged}}$ indicates the merged model. This work targets data-free arithmetic-based operators, which instantiate $\mathcal{F}$ as non-iterative per-parameter rules
applied to $\{\tau_i\}$. 

CUSD does not change this formulation; instead, it pre-processes the task vectors before $\mathcal{F}$ is applied.
These operators act on scalar entries of $\{\tau_i\}$ and therefore do not expose the matrix-level directional structure of a task update.
CUSD instead projects each matrix-structured layer into a shared directional basis, where directions with comparable cross-task energy can be separated from directions dominated by one or a few tasks (Figure~\ref{fig:overview}).
For each retained layer~$\ell$ with matrix-structured weights, we denote the corresponding per-task matrices by $A_1^{(\ell)}, \ldots, A_K^{(\ell)} \in \mathbb{R}^{m \times n}$; all subsequent operations are applied independently per layer, and we suppress~$\ell$ for clarity.

\subsection{HO-GSVD-Derived Shared Basis}
\label{sec:hogsvd}

A standard SVD decomposes a single matrix as $A=U\Sigma V^\top$.
  Applied separately to $K$ task matrices, it produces unrelated right
  bases. Classical HO-GSVD instead seeks a shared right factor and is
  commonly written as
  \[
    A_i = U_i \Sigma_i V^\top,\qquad i\in[K],
  \]
  which motivates comparing task matrices in a common right basis.
  CUSD uses the regularized and symmetrized Gram-quotient construction
  in Appendix~\ref{app:hogsvd_derivation} to obtain an orthonormal
  shared basis $V=[v_1,\ldots,v_n]$, which is required for Eq.~\ref{eq:proj}.

  For each shared direction, CUSD records the directional energy
  $\sigma_{ik}=\|A_i v_k\|_2$ and collects these values into
  $\Sigma=[\sigma_{ik}]\in\mathbb{R}^{K\times n}$.

 Because all tasks are evaluated along the same directions, these energies are directly comparable across tasks. This shared-basis construction is computed once for each retained matrix-structured layer and reused across merging methods.

\subsection{The CUSD Algorithm}
\label{sec:classify}

\paragraph{Std-based dispersion score.}
Given the matrix $\Sigma$, CUSD measures the cross-task imbalance of each shared direction. For direction~$k$, the std-based dispersion score is
\begin{equation}
  d_k
  =
  \operatorname{Std}_{i \in [K]}\!\left(\tilde{\sigma}_{ik}\right),
  \qquad
  \tilde{\sigma}_{ik}
  =
  \begin{cases}
    \ \ \ \ \ \ \sigma_{ik}, & \text{absolute scoring},\\
    \frac{\sigma_{ik}}{K^{-1}\sum_i \sigma_{ik}}, & \text{relative scoring}.
  \end{cases}
  \label{eq:dispersion_score}
\end{equation}
Directions with comparable cross-task energy receive small $d_k$, whereas directions dominated by one or a few tasks receive large $d_k$. Relative scoring removes the per-direction mean energy scale and measures relative cross-task imbalance. We compare the two scoring choices empirically in Section~\ref{sec:ablation}.

\paragraph{KDE.}
The threshold separating common from unique is determined automatically from the distribution of $\{d_k\}_{k=1}^n$ via kernel density estimation (KDE) with Scott's bandwidth~\citep{scott2015multivariate} (full details appear in Appendix~\ref{app:kde}). Let $\delta$ denote the detected valley. CUSD then sets
\begin{equation}
  \mathcal{C} = \{k : d_k < \delta\},
  \qquad
  \mathcal{U} = \{k : d_k \ge \delta\}.
  \label{eq:kde_partition}
\end{equation}

\paragraph{Subspace merging.}
Given the partition $(\mathcal{C}, \mathcal{U})$, let $V_\mathcal{C}$ and $V_\mathcal{U}$ denote the columns of $V$ indexed by $\mathcal{C}$ and $\mathcal{U}$, respectively.
Define
\begin{equation}
  P_\mathcal{C} = V_\mathcal{C} V_\mathcal{C}^\top,\qquad
  P_\mathcal{U} = V_\mathcal{U} V_\mathcal{U}^\top = I - P_\mathcal{C},
  \qquad
  C_i = A_i P_\mathcal{C},\qquad
  R_i = A_i P_\mathcal{U}.
  \label{eq:proj}
\end{equation}
Because $P_\mathcal{C}$ and $P_\mathcal{U}$ are complementary orthogonal projectors, this gives an exact orthogonal decomposition: $C_i + R_i = A_i$ and $\langle C_i, R_i \rangle_F = 0$; proof in Appendix~\ref{app:lossless}.
The common component $C_i$ contains directions that are consistently emphasized across tasks, whereas the unique residual $R_i$ contains directions whose energy is concentrated in one or a few tasks and therefore reflects task-specific variation.
CUSD therefore applies the merging operator $\mathcal{F}$ only to $\{C_i\}$, while the unique residuals $\{R_i\}$ are summed and scaled by $\beta \in [0,1]$. The resulting layer-wise merged task vector is
\begin{equation}
  \tau_m
  =
  \mathcal{F}\!\left(C_1, \ldots, C_K\right)
  +
  \beta \sum\nolimits_{i=1}^{K} R_i .
  \label{eq:cusd_layer_merge}
\end{equation}
Setting $\beta = 0$ discards all unique directions; $\beta = 1$ retains them in full.
Algorithm~\ref{alg:cusd} summarizes the complete per-layer CUSD procedure. 
The decomposition (lines~2--5) depends only on the task vectors and the classification rule, not on $\mathcal{F}$.
This pre-processing is computed once and reused with any arithmetic-based merging method.

\begin{algorithm}[t]
\caption{CUSD: Common--Unique Subspace Decomposition}
\label{alg:cusd}
\begin{algorithmic}[1]
\REQUIRE $\tau_1, \ldots, \tau_K$ (task vectors); $\mathcal{F}(\cdot)$ (any arithmetic-based method); $\beta \in [0,1]$
\ENSURE $\theta_{\mathrm{merged}}$ (merged model)
    \FOR{each retained layer $\ell$ with matrix-structured weights}
    \STATE $V,\, \Sigma \gets \text{SharedBasis} \ \!\bigl(A_1^{(\ell)}, \ldots, A_K^{(\ell)}\bigr)$ \COMMENT{Alg.~\ref{alg:hogsvd}}
    \STATE $\mathcal{C},\, \mathcal{U} \gets \text{Classify}(\Sigma)$ \COMMENT{std-based dispersion + KDE; \S\ref{sec:classify}}
    \STATE $P_\mathcal{C} \gets V_\mathcal{C} \, V_\mathcal{C}^\top$;\quad $P_\mathcal{U} \gets I - P_\mathcal{C}$ \COMMENT{Orthogonal projectors}

    \FOR{$i = 1, \ldots, K$} \STATE $C_i \gets A_i \, P_\mathcal{C}$;\quad $R_i \gets A_i \, P_\mathcal{U}$ \quad \COMMENT{Prop.~\ref{prop:lossless}} 
    \ENDFOR
    \STATE $\displaystyle \tau_m^{(\ell)} \gets \mathcal{F}(C_1, \ldots, C_K) \;+\; \beta \sum\nolimits_{i=1}^{K} R_i$ \COMMENT{Eq.~\ref{eq:cusd_layer_merge}}
\ENDFOR
\RETURN $\theta_{\mathrm{merged}}$ $\gets$ $\theta_{\mathrm{base}} +  \tau_m$
\end{algorithmic}
\end{algorithm}

Overall, CUSD augments an arithmetic-based merging operator with three components: a shared-basis decomposition, a std-based common--unique partition, and a $\beta$-controlled unique residual. 

%% file: sections/experiment.tex
\section{Experiments}
\label{sec:experiments}

\subsection{Experimental Setup}
\label{sec:setup}

 \paragraph{Models and tasks.}
  We evaluate CUSD on 11 base models covering vision encoders, text encoders, and LLMs: CLIP ViT-B/32, ViT-B/16, and ViT-L/14~\citep{radford2021learning,dosovitskiy2020image};
  RoBERTa-Base and RoBERTa-Large~\citep{liu2019roberta}; and MergeBench Llama and Gemma models~\citep{he2025mergebench,grattafiori2024llama,team2024gemma}. The evaluation follows prior model-
  merging benchmarks on eight image-classification datasets~\citep{yang2023adamerging,cheng2025whoever}, eight GLUE NLU tasks~\citep{wang2018glue,cheng2025whoever}, and MergeBench code, math,
  and multilingual LLM tasks~\citep{he2025mergebench}. Full model and dataset details are provided in Appendix~\ref{app:setup}.

  \paragraph{Baselines and implementation.}
  We pair CUSD with four representative data-free arithmetic-based merging methods: Task Arithmetic~\citep{ilharco2022editing}, TIES-Merging~\citep{yadav2023ties}, DARE~\citep{yu2024language},
  and Consensus Merging~\citep{wang2024localizing}. For each method, we report the plain merge and the corresponding CUSD-preprocessed merge; TSV-Merge~\citep{gargiulo2025task} and Iso-
  Merging~\citep{marczak2025no} are included in the subspace-based comparison. We also evaluate CUSD preprocessing with
WUDI-Merging~\citep{cheng2025whoever}
and ACE-Merging~\citep{xu2026ace}. CUSD is applied to matrix-structured linear-layer task vectors. Unless otherwise specified, baseline hyperparameters follow the
  original papers or MergeBench defaults and are kept fixed between plain and CUSD-preprocessed runs. Additional implementation details are provided in Appendix~\ref{app:setup}, and Section~\ref{sec:ablation} reports the scoring mode and $\beta$ sensitivity used in the main experiments.

\subsection{Main Results}
\label{sec:main}

 \paragraph{Overall performance.}
  Table~\ref{tab:overall_performance} reports average normalized performance across all model families. Parenthesized values denote changes from the corresponding plain merge, and boldface
  marks the best score in each model column. CUSD improves $42$ of the $44$ model--operator combinations, with one tie and one small decrease. The largest gain is $22.5$ normalized points on
  RoBERTa-Large, and all ViT and RoBERTa settings improve. On LLMs, CUSD improves $22$ of $24$ settings. Detailed task-level results in Appendix~\ref{app:main_results} show that these gains often reflect an improved multi-task trade-off: small drops on already
  strong tasks are accompanied by larger gains on other tasks, consistent with separating unique directions before arithmetic aggregation.

\begin{table}[htbp]
\caption{\textbf{Overall performance across all models.} We report average normalized scores.}
\label{tab:overall_performance}
\centering
\scriptsize
\setlength{\tabcolsep}{2.0pt}
\resizebox{\linewidth}{!}{%
\begin{tabular}{lc ccccccccccc}
\toprule
\textbf{Method} & with \textbf{CUSD} & \textbf{ViT-B/32} & \textbf{ViT-B/16} & \textbf{ViT-L/14} & \textbf{RoBERTa-B} & \textbf{RoBERTa-L} & \textbf{Llama-3B} & \textbf{Llama-8B} & \textbf{Llama-3B-Inst} & \textbf{Llama-8B-Inst} & \textbf{Gemma-2b} & \textbf{Gemma-9b} \\
\midrule
Fine-tuned & --- & \resscore{100.0} & \resscore{100.0} & \resscore{100.0} & \resscore{100.0} & \resscore{100.0} & \resscore{100.0} & \resscore{100.0} & \resscore{100.0} & \resscore{100.0} & \resscore{100.0} & \resscore{100.0} \\
\midrule
\multirow{2}{*}{Task Arithmetic}
 & \resx & \resscore{71.8} & \resscore{76.8} & \resscore{87.6} & \resscore{53.5} & \resscore{58.3} & \resscore{69.6} & \resscore{93.7} & \resscore{73.5} & \resscore{82.5} & \resscore{94.2} & \resscore{92.0} \\
 & \rescheck & \resup{85.0}{13.2} & \resup{88.9}{12.1} & \resup{93.0}{5.4} & \resup{68.7}{15.1} & \resup{\textbf{70.0}}{11.7} & \ressame{69.6}{0.0} & \resup{\textbf{94.1}}{0.4} & \resup{75.8}{2.3} & \resup{86.5}{4.0} & \resup{95.7}{1.5} & \resup{94.8}{2.8} \\
\midrule
\multirow{2}{*}{TIES-Merging}
 & \resx & \resscore{74.0} & \resscore{78.5} & \resscore{88.5} & \resscore{51.1} & \resscore{47.3} & \resscore{69.2} & \resscore{93.7} & \resscore{72.7} & \resscore{82.6} & \resscore{94.1} & \resscore{91.1} \\
 & \rescheck & \resup{\textbf{87.0}}{13.0} & \resup{\textbf{90.3}}{11.8} & \resup{\textbf{94.8}}{6.3} & \resup{\textbf{69.3}}{18.2} & \resup{69.8}{22.5} & \resup{69.7}{0.5} & \resup{\textbf{94.1}}{0.4} & \resup{75.5}{2.8} & \resup{86.2}{3.6} & \resup{\textbf{97.3}}{3.2} & \resup{95.3}{4.2} \\
\midrule
\multirow{2}{*}{DARE}
 & \resx & \resscore{71.7} & \resscore{76.6} & \resscore{87.5} & \resscore{54.0} & \resscore{58.8} & \resscore{68.9} & \resscore{\textbf{94.1}} & \resscore{73.4} & \resscore{83.1} & \resscore{94.1} & \resscore{91.2} \\
 & \rescheck & \resup{84.8}{13.2} & \resup{88.8}{12.2} & \resup{92.9}{5.4} & \resup{68.3}{14.3} & \resup{69.5}{10.7} & \resup{70.3}{1.4} & \resdown{93.8}{0.3} & \resup{75.4}{2.0} & \resup{86.4}{3.3} & \resup{96.2}{2.1} & \resup{94.8}{3.6} \\
\midrule
\multirow{2}{*}{Consensus}
 & \resx & \resscore{73.2} & \resscore{78.0} & \resscore{88.0} & \resscore{55.2} & \resscore{61.9} & \resscore{67.4} & \resscore{93.8} & \resscore{76.2} & \resscore{86.2} & \resscore{88.8} & \resscore{94.6} \\
 & \rescheck & \resup{85.0}{11.8} & \resup{88.8}{10.8} & \resup{92.8}{4.9} & \resup{67.3}{12.1} & \resup{\textbf{70.0}}{8.1} & \resup{\textbf{70.6}}{3.2} & \resup{93.9}{0.1} & \resup{\textbf{76.6}}{0.4} & \resup{\textbf{87.5}}{1.3} & \resup{93.1}{4.3} & \resup{\textbf{95.5}}{0.9} \\
\bottomrule
\end{tabular}
}
\end{table}

 \paragraph{Results on encoder models.}
  On vision encoders, CUSD improves all four operators on all three ViT models, with average gains of about $12$ points on ViT-B/32 and ViT-B/16.
  Appendix Tables~\ref{tab:vit_l14}--\ref{tab:roberta_base} show a trade-off pattern: with TIES-Merging, CUSD reduces the already strong MNIST score by only $0.1$ points while improving three weaker tasks by $8.5$--$10.7$ points.
On text encoders, CUSD improves every operator on both RoBERTa models, and the detailed GLUE tables show that STS-B changes from negative plain-merge scores to positive CUSD scores across operators. These trends support the view that common--unique separation improves the multi-task balance in encoder merging.

  \paragraph{Results on LLMs.}
  On generative LLMs, CUSD improves most settings. The strongest gains appear on the Llama Instruct and Gemma models, while Llama-3.1-8B leaves little headroom because plain merges already reach normalized averages around $94$; the only negative LLM change is a $0.3$-point decrease for DARE on this model. The role of partial unique-
  residual scaling in LLMs is examined in the $\beta$ ablation (Section~\ref{sec:ablation}) and analysis (Section~\ref{sec:analysis}). Full task-level LLM results are provided in Appendix
  Tables~\ref{tab:llm_llama32_3b}--\ref{tab:llm_gemma2_9b}.

\paragraph{Comparison with other merging methods.}
Table~\ref{tab:subspace_comparison} compares CUSD+TA with
TSV-M, Iso-C, and Iso-CTS across four model families.
For LLMs, the SVD baselines report the best-observed scores
from post-hoc scaling searches on the test sets.
Iso-CTS performs best on ViT-B/16 and Iso-C on Gemma-2-2b,
whereas CUSD+TA achieves the highest scores on
RoBERTa-Base and Llama-3.2-3B.

We further evaluate CUSD preprocessing with WUDI-Merging
and ACE-Merging (Table~\ref{tab:wudi_ace_cusd}).
CUSD improves WUDI in three of four settings and ACE
on both LLMs, while encoder results are mixed.

\begin{table}[t]
\centering
\begin{minipage}[t]{0.49\linewidth}
\centering
\caption{Comparison with subspace-based merging methods.
Average normalized scores are reported.}
\label{tab:subspace_comparison}
\small
\setlength{\tabcolsep}{2.5pt}
\resizebox{\linewidth}{!}{%
\begin{tabular}{lcccc}
\toprule
Method
& \makecell{ViT\\B/16}
& \makecell{RoBERTa\\Base}
& \makecell{Llama\\3.2-3B}
& \makecell{Gemma\\2-2b} \\
\midrule
Fine-tuned & 100.0 & 100.0 & 100.0 & 100.0 \\
\midrule
TSV-M   & 93.4 & 45.9 & 64.5 & 96.1 \\
Iso-C   & 94.5 & 42.9 & 65.7 & \textbf{97.7} \\
Iso-CTS & \textbf{95.6} & 42.8 & 64.5 & 97.2 \\
CUSD+TA & 88.9 & \textbf{68.7} & \textbf{69.6} & 95.7 \\
\bottomrule
\end{tabular}%
}
\end{minipage}
\hfill
\begin{minipage}[t]{0.49\linewidth}
\centering
\caption{Effect of CUSD preprocessing on WUDI-Merging
and ACE-Merging. Average normalized scores are reported.}
\label{tab:wudi_ace_cusd}
\small
\setlength{\tabcolsep}{2.5pt}
\resizebox{\linewidth}{!}{%
\begin{tabular}{lcccc}
\toprule
Method
& \makecell{ViT\\B/16}
& \makecell{RoBERTa\\Base}
& \makecell{Llama\\3.2-3B}
& \makecell{Gemma\\2-2b} \\
\midrule
Fine-tuned & 100.0 & 100.0 & 100.0 & 100.0 \\
\midrule
WUDI      & \textbf{95.6} & 69.3 & 65.6 & 93.6 \\
CUSD+WUDI & 94.0 & \textbf{80.7}
          & \textbf{66.9} & \textbf{94.9} \\
\midrule
ACE      & \textbf{96.4} & \textbf{85.6} & 61.9 & 93.7 \\
CUSD+ACE & 94.7 & 81.1 & \textbf{63.3} & \textbf{98.6} \\
\bottomrule
\end{tabular}%
}
\end{minipage}
\end{table}

\subsection{Ablation Study}
\label{sec:ablation}

\paragraph{Dispersion scoring choice.}
We compare absolute and relative scoring
(Eq.~\ref{eq:dispersion_score}).
Relative scoring performs better in six of the eight
LLM model--operator settings, while absolute scoring
performs better with Task Arithmetic on both models
(Table~\ref{tab:llm_scoring_mode}). These results support relative scoring as the preferred choice for the evaluated LLM settings.
Section~\ref{sec:analysis} examines how cross-task directional-energy patterns motivate the choice between absolute and relative scoring.

\begin{table}[htbp]
\caption{Absolute vs. relative dispersion scoring on decoder LLMs. We report normalized average performance; green values denote gains over the corresponding plain merge.}
\label{tab:llm_scoring_mode}
\centering
\scriptsize
\setlength{\tabcolsep}{2.5pt}
\resizebox{\linewidth}{!}{%
\begin{tabular}{lcccccccc}
\toprule
\multirow{2}{*}{\textbf{Method}} &
\multicolumn{4}{c}{\textbf{Llama-3.2-3B}} &
\multicolumn{4}{c}{\textbf{Gemma-2-2b}} \\
\cmidrule(lr){2-5}\cmidrule(lr){6-9}
& \textbf{w/o CUSD} & \textbf{CUSD ($s_{\mathrm{abs}}$)} & \textbf{CUSD ($s_{\mathrm{rel}}$)} & \textbf{$s_{\mathrm{rel}}$ - $s_{\mathrm{abs}}$}
& \textbf{w/o CUSD} & \textbf{CUSD ($s_{\mathrm{abs}}$)} & \textbf{CUSD ($s_{\mathrm{rel}}$)} & \textbf{$s_{\mathrm{rel}}$ - $s_{\mathrm{abs}}$} \\
\midrule
Task Arithmetic & 69.6 & \textbf{70.0}{\scriptsize\textcolor{ForestGreen}{(+0.4)}} & 69.6{\scriptsize\textcolor{ForestGreen}{(+0.0)}} & -0.4 & 94.2 & \textbf{96.9}{\scriptsize\textcolor{ForestGreen}{(+2.7)}} & 95.7{\scriptsize\textcolor{ForestGreen}{(+1.5)}} & -1.2 \\
TIES-Merging & 69.2 & 69.5{\scriptsize\textcolor{ForestGreen}{(+0.3)}} & \textbf{69.7}{\scriptsize\textcolor{ForestGreen}{(+0.5)}} & 0.2 & 94.1 & 95.9{\scriptsize\textcolor{ForestGreen}{(+1.8)}} & \textbf{97.3}{\scriptsize\textcolor{ForestGreen}{(+3.2)}} & 1.4 \\
DARE & 68.9 & 69.6{\scriptsize\textcolor{ForestGreen}{(+0.7)}} & \textbf{70.3}{\scriptsize\textcolor{ForestGreen}{(+1.4)}} & 0.7 & 94.1 & 95.8{\scriptsize\textcolor{ForestGreen}{(+1.7)}} & \textbf{96.2}{\scriptsize\textcolor{ForestGreen}{(+2.1)}} & 0.4 \\
Consensus & 67.4 & 70.0{\scriptsize\textcolor{ForestGreen}{(+2.6)}} & \textbf{70.6}{\scriptsize\textcolor{ForestGreen}{(+3.2)}} & 0.6 & 88.8 & 91.7{\scriptsize\textcolor{ForestGreen}{(+2.9)}} & \textbf{93.1}{\scriptsize\textcolor{ForestGreen}{(+4.3)}} & 1.4 \\
\bottomrule
\end{tabular}
}
\end{table}

\paragraph{Direction selection.}
To examine whether the choice of unique directions matters, we compare CUSD with Random, Reversed (lowest dispersion),
and Top-energy selection. In the matched controls, each
method selects the same number of unique directions as CUSD
in every layer. We also test Top-energy with fixed selection
fractions. CUSD outperforms all controls on both backbones
(Table~\ref{tab:direction_selection}); Top-energy is more
competitive than Random and Reversed, consistent with the
energy concentration of CUSD's unique directions.

\begin{table}[t]
\centering
\caption{Direction-selection ablation for CUSD+TA.
Average accuracy (\%) is reported; Random shows mean
$\pm$ standard deviation over three seeds.}
\label{tab:direction_selection}
\small
\setlength{\tabcolsep}{3pt}
\resizebox{\linewidth}{!}{%
\begin{tabular}{lccccccc}
\toprule
Model & Random & Reversed
& \makecell{Top-energy\\(matched)}
& \makecell{Top-energy\\(0.2\%)}
& \makecell{Top-energy\\(0.5\%)}
& \makecell{Top-energy\\(1.0\%)}
& CUSD \\
\midrule
ViT-B/32 & $59.20 \pm 0.08$ & 59.57 & 73.78
& 74.75 & 73.68 & 70.97 & \textbf{76.48} \\
ViT-B/16 & $65.64 \pm 1.17$ & 65.09 & 81.07
& 80.84 & 80.34 & 78.97 & \textbf{81.74} \\
\bottomrule
\end{tabular}%
}
\end{table}

\paragraph{Common--unique merging design.}
We apply $\mathcal{F}$ to common components only
(Common, Eq.~\ref{eq:cusd_layer_merge}), unique components
only (Unique), or both separately (Both).
Common performs best in all four reported settings
(Table~\ref{tab:merge_design}), supporting common aggregation
with separately controlled unique-residual retention.

\begin{figure}[H]
  \begin{minipage}{0.45\textwidth}
  \includegraphics[width=0.8\textwidth]{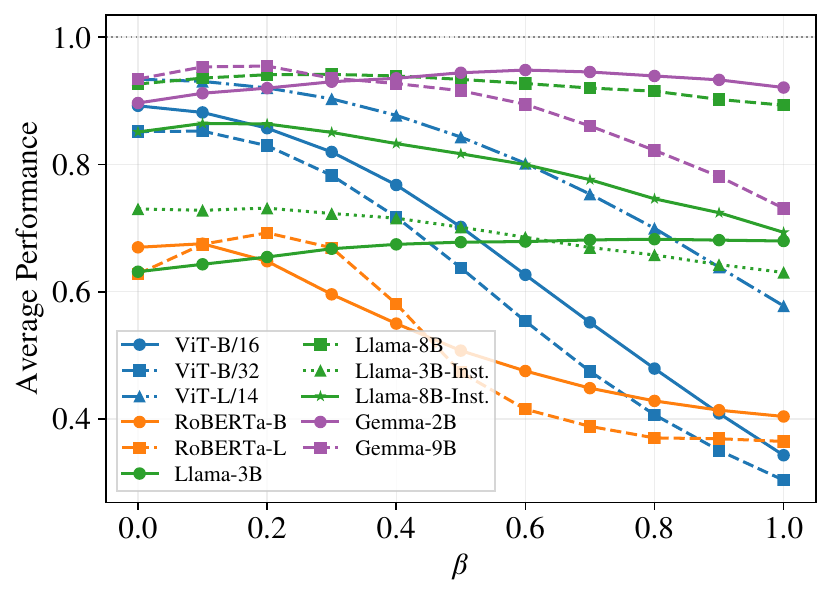}

\caption{Sensitivity to the unique-residual retention coefficient $\beta$ across all 11 models.}
\label{fig:beta_sweep}
 
  \end{minipage}
  \hfill
  \begin{minipage}{0.45\textwidth}
\includegraphics[width=\textwidth]{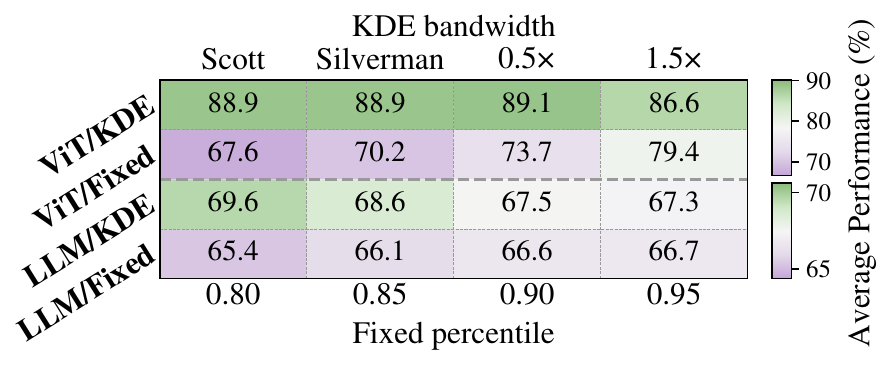}

\caption{Average performance under KDE-based thresholds and fixed percentile thresholds. For KDE rows, columns denote bandwidth choices; for fixed-threshold rows, columns denote percentile cutoffs.}
\label{fig:kde_threshold_heatmap}

  \end{minipage}
\end{figure}

\paragraph{Unique-residual coefficient.}
\label{para:beta_sweep}
The coefficient $\beta$ controls the retention strength of the unique residual. We sweep $\beta \in \{0.0,0.1,\ldots,1.0\}$ to examine how sensitive CUSD is to this residual contribution. Figure~\ref{fig:beta_sweep} shows that encoders favor small $\beta$, whereas decoder LLMs benefit from moderate residual retention, typically around $\beta\in[0.2,0.5]$. Section~\ref{para:unique} analyzes this difference. Since per-operator sweeps in Figure~\ref{fig:beta_per_method_consistency} show consistent trends across downstream merge operators, Figure~\ref{fig:beta_sweep} reports the average over the four methods.

We further examine how $\beta$ interacts with the downstream hyperparameter $\lambda$, using Task Arithmetic as a case study.
For CUSD+TA, $\tau_m=\lambda\sum_i C_i+\beta\sum_i R_i$,
which reduces to plain TA when $\beta=\lambda$.
Table~\ref{tab:ta_joint_sweep} reports a post-hoc joint sweep
on ViT-B/32. CUSD+TA outperforms plain TA at the same $\lambda$
in every tested configuration with $\lambda\geq0.3$ and
$\beta<\lambda$. These results show that reducing the
unique-residual contribution relative to the common component
can improve merging performance in this setting.

\paragraph{Adaptive KDE thresholding.}

 CUSD uses KDE valley detection to determine the common--unique partition from the dispersion-score distribution, avoiding a fixed global cutoff. Figure~\ref{fig:kde_threshold_heatmap} compares this adaptive threshold with fixed percentile thresholds. KDE thresholding consistently outperforms fixed thresholds on both ViT and LLM settings, with the largest gap on ViT where fixed percentiles are sensitive to the chosen cutoff. The result is also stable across standard KDE bandwidth choices: Scott's and Silverman's rules give nearly identical performance, and moderate bandwidth changes preserve the overall advantage of KDE. This supports using a data-driven layer-adaptive partition rather than tuning a fixed percentile threshold.

\begin{table}[htbp]
\centering
\begin{minipage}[t]{0.43\linewidth}
\centering
\caption{Merging-design ablation.
Average normalized scores are reported;
Common denotes CUSD.}
\label{tab:merge_design}
\small
\setlength{\tabcolsep}{2.5pt}
\resizebox{\linewidth}{!}{%
\begin{tabular}{llccc}
\toprule
Model & Operator & Common & Unique & Both \\
\midrule
ViT-B/16 & TIES & \textbf{90.3} & 73.0 & 72.4 \\
ViT-B/16 & DARE & \textbf{88.8} & 76.4 & 76.3 \\
\midrule
Gemma-2-2b & TIES & \textbf{97.3} & 94.2 & 95.2 \\
Gemma-2-2b & DARE & \textbf{96.2} & 93.6 & 94.2 \\
\bottomrule
\end{tabular}%
}
\end{minipage}
\hfill
\begin{minipage}[t]{0.54\linewidth}
\centering
\caption{One-time CUSD decomposition cost.}
\label{tab:computation_resource}
\small
\setlength{\tabcolsep}{2.5pt}
\resizebox{\linewidth}{!}{%
\begin{tabular}{lrrrrr}
\toprule
Model & Tasks & Layers
& \makecell{Total\\time}
& \makecell{Time\\per layer}
& \makecell{GPU\\peak} \\
\midrule
ViT-B/32     & 8 & 48  & 16.5s  & 0.343s & 1.1 GB \\
ViT-B/16     & 8 & 48  & 16.3s  & 0.340s & 1.1 GB \\
ViT-L/14     & 8 & 96  & 48.5s  & 0.505s & 1.9 GB \\
Llama-3B & 3 & 196 & 67.5s  & 0.345s & 5.0 GB \\
Llama-8B & 3 & 224 & 217.2s & 0.970s & 15.4 GB \\
\bottomrule
\end{tabular}%
}
\end{minipage}

\end{table}

\subsection{Analysis}
\label{sec:analysis}

\begin{figure}[t]
  \begin{minipage}{0.45\textwidth}

\includegraphics[width=\textwidth]{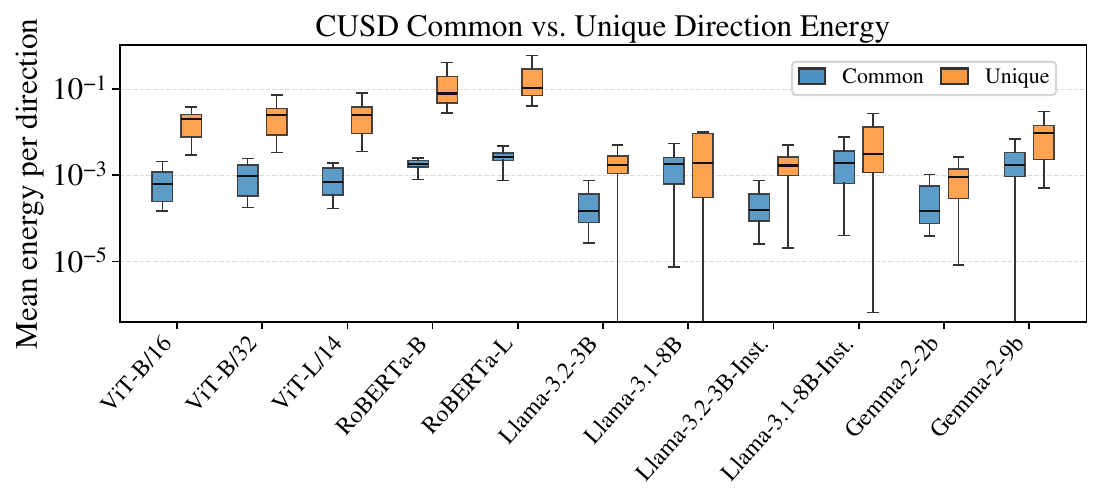}
\caption{Layer-wise mean energy per direction
in common and unique subspaces.}
\label{fig:common-unique-energy-boxplot}
 
  \end{minipage}
  \hfill
  \begin{minipage}{0.45\textwidth}
\includegraphics[width=\textwidth]{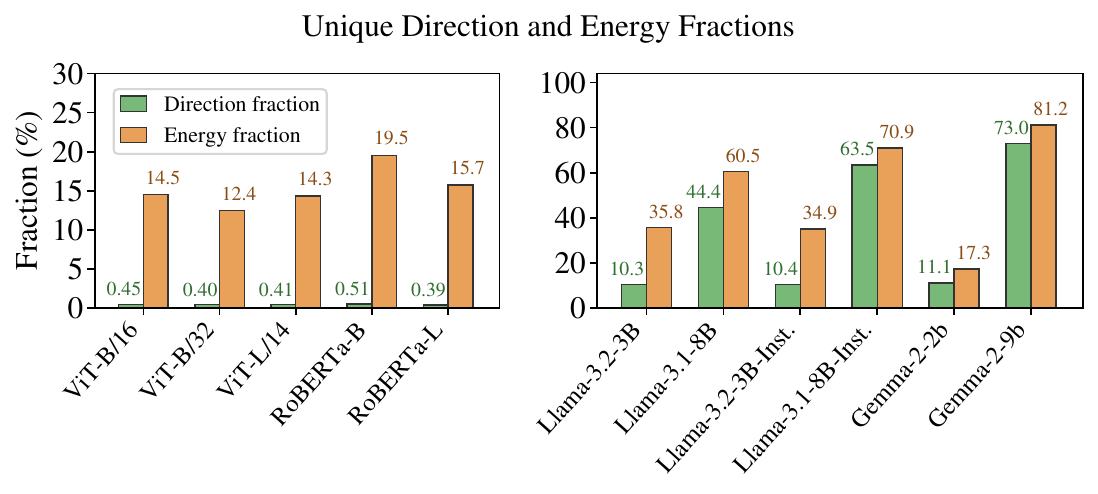}
\caption{Fractions of directions and total energy
assigned to the unique subspace.}
\label{fig:unique-energy-fraction}

  \end{minipage}
\end{figure}

\paragraph{What CUSD separates.}
The common--unique partition gives an operational view of the two-regime structure observed in Figure~\ref{fig:motivation} and further visualized in Appendix~\ref{app:directional_structure}: directions assigned to the common subspace form a broad low-dispersion bulk, whereas directions assigned to the unique residual are often fewer but carry substantially higher per-direction energy. Figure~\ref{fig:common-unique-energy-boxplot} confirms this pattern across the evaluated model families. Unique directions consistently have higher per-direction energy than common directions, especially in encoder models. This shows that CUSD does not merely discard weak or low-energy directions; it separates an energy-concentrated residual subspace whose directions are unevenly supported across tasks. As summarized in Figure~\ref{fig:unique-energy-fraction}, this concentration is particularly strong in encoders: fewer than $1\%$ of directions account for roughly $12$--$20\%$ of the directional energy.

\paragraph{Understanding dispersion scoring.}

\begin{figure}[h]
\centering
\includegraphics[width=\linewidth]{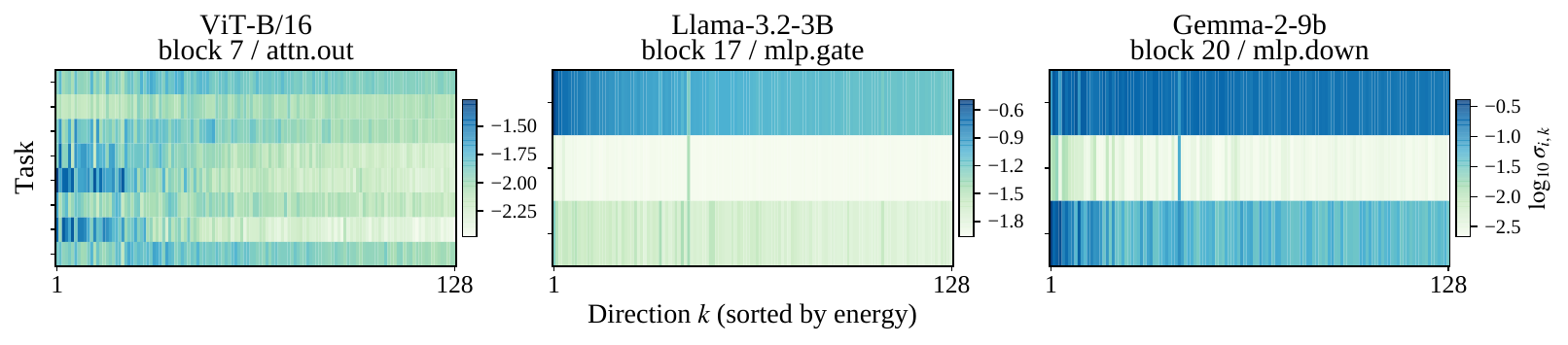}
\caption{Per-task directional singular values for representative ViT, Llama, and Gemma layers. Each panel shows the HO-GSVD singular values $\sigma_{ik}$ of task vectors along the top shared directions.}
\label{fig:sigma_heatmap_vit_llama}
\end{figure}

\begin{figure}[t]
\centering
\begin{subfigure}[t]{0.49\linewidth}
  \centering
  \includegraphics[width=0.8\linewidth]{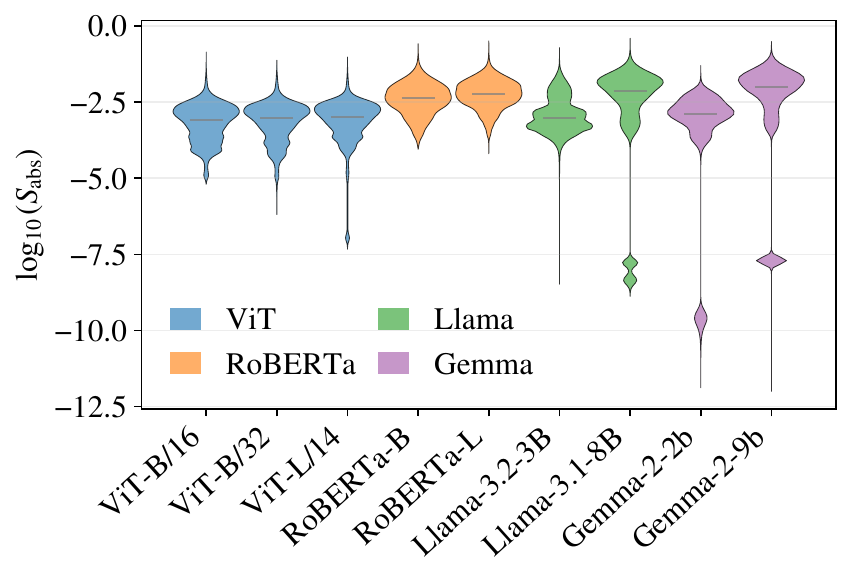}
  \caption{Absolute dispersion across all blocks/directions.}
  \label{fig:sabs_violin}
\end{subfigure}
\hfill
\begin{subfigure}[t]{0.49\linewidth}
  \centering
  \includegraphics[width=0.8\linewidth]{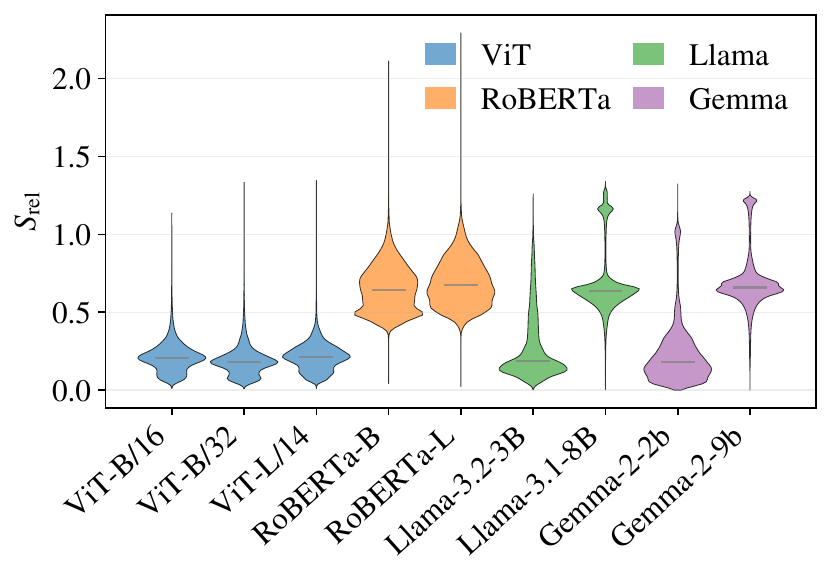}
  \caption{Relative dispersion across all blocks/directions.}
  \label{fig:srel_violin}
\end{subfigure}
\caption{Distribution of dispersion scores under absolute and relative scoring. Mean normalization changes the dispersion distribution, exposing scale-suppressed task imbalance in the evaluated LLM settings while amplifying low-energy relative fluctuations in the evaluated encoder settings.}
\label{fig:dispersion_violin}
\end{figure}

For decoder LLMs, small absolute dispersion does not necessarily mean that a direction is uniformly balanced across tasks. If such directions were simply shared evenly by all tasks, mean normalization would keep their relative dispersion small. Instead, Figure~\ref{fig:dispersion_violin} shows that the low-dispersion mass under $s_{\mathrm{abs}}$ is substantially reduced under $s_{\mathrm{rel}}$. Together with the task-imbalanced energy patterns in Figure~\ref{fig:sigma_heatmap_vit_llama}, this indicates that many low-absolute-dispersion directions are low or moderate energy directions that are nevertheless dominated by specific tasks. Relative scoring removes this raw-scale suppression and exposes these directions as high-dispersion.

The encoder case is different. In ViT, the heatmap in Figure~\ref{fig:sigma_heatmap_vit_llama} shows more balanced directional energies across tasks, except for a small high-energy region with visible task dominance. In this regime, $s_{\mathrm{abs}}$ preserves the energy-concentration signal, whereas $s_{\mathrm{rel}}$ gives low-energy directions the same scale as high-energy ones and can amplify small relative fluctuations. The ViT distributions in Figure~\ref{fig:dispersion_violin} are consistent with this effect: relative scoring introduces a longer high-dispersion tail despite the comparatively balanced heatmap. This can weaken the separation between high-energy task-dominated directions and low-energy variations.  This supports our main-experiment choice of using $s_{\mathrm{abs}}$ for the evaluated encoder settings and $s_{\mathrm{rel}}$ for the evaluated LLM settings. Appendix Figures~\ref{fig:sigma_heatmap_vit_b16_appendix} and~\ref{fig:sigma_heatmap_llama_3b_appendix} provide additional layer heatmaps.

\paragraph{Unique residual scale across model families.} 
\label{para:unique}
Figure~\ref{fig:unique-energy-fraction} shows that the unique subspace differs across the evaluated model families and task groups. In encoders, the unique subspace is extremely sparse: it contains fewer than $1\%$ of directions but carries roughly $12$--$20\%$ of the directional energy. This creates a concentrated residual component, so increasing $\beta$ can quickly reintroduce task-specific interference; this explains why encoder performance in Figure~\ref{fig:beta_sweep} is sensitive to $\beta$ and why $\beta=0$ is a stable default. In decoder LLMs, the unique subspace is much larger and carries a substantial fraction of the energy, from $17.3\%$ to $81.2\%$. Removing this broader residual component entirely can discard task-specific generative information, while moderate residual scaling preserves part of it without directly merging it into the common component. This explains why LLMs prefer nonzero $\beta$ and remain less sensitive over a moderate range of $\beta$ values.

\paragraph{Interference reduction.}

To examine whether CUSD reduces interference inside the merged update, we measure a normalized interference proxy (NI), defined in Appendix~\ref{app:interference_analysis}. Figure~\ref{fig:interference_dare_vit} compares DARE and CUSD-DARE on the three ViT backbones. CUSD consistently lowers NI across blocks, with average reductions of $10.0\%$, $9.0\%$, and $10.2\%$ on ViT-B/32, ViT-B/16, and ViT-L/14, respectively. This supports the proposed mechanism: separating common and unique directions before merging reduces the parameter-space interference that arises when arithmetic rules are applied directly to task-dominated directions. The same trend holds for other downstream merge operators, as reported in Appendix~\ref{app:interference_analysis}.

\begin{figure}[h]
\centering
\begin{subfigure}[t]{0.32\linewidth}
  \centering
  \includegraphics[width=\linewidth]{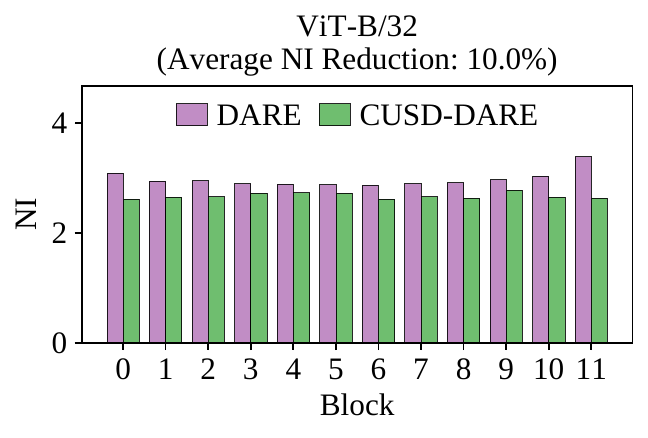}
  \caption{ViT-B/32.}
  \label{fig:interference_dare_vit_b32}
\end{subfigure}
\hfill
\begin{subfigure}[t]{0.32\linewidth}
  \centering
  \includegraphics[width=\linewidth]{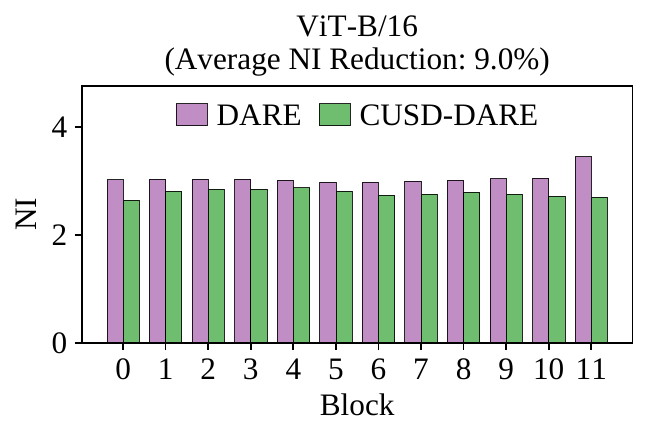}
  \caption{ViT-B/16.}
  \label{fig:interference_dare_vit_b16}
\end{subfigure}
\hfill
\begin{subfigure}[t]{0.32\linewidth}
  \centering
  \includegraphics[width=\linewidth]{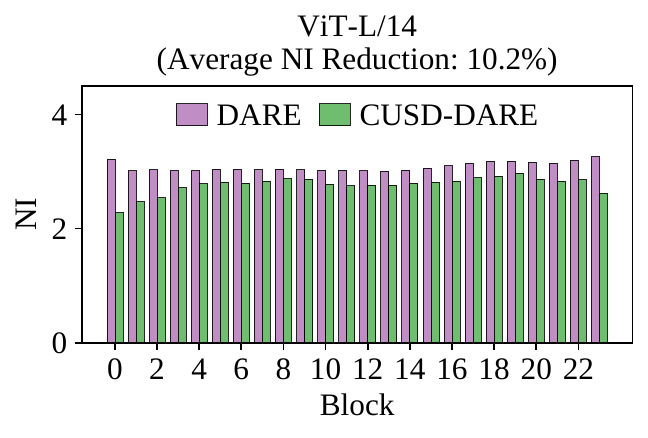}
  \caption{ViT-L/14.}
  \label{fig:interference_dare_vit_l14}
\end{subfigure}
\caption{\textbf{Per-block interference reduction.} DARE and CUSD-DARE are compared on ViT models using the normalized interference proxy (NI); lower values indicate less deviation from the task-specific updates.}
\label{fig:interference_dare_vit}
\end{figure}

\paragraph{Computation cost.}
 CUSD adds a one-time shared basis decomposition whose common components and unique residuals are reused across merge operators. Table~\ref{tab:computation_resource} shows that ViT decompositions
  take $16.3$--$48.5$s with at most $1.9$GB peak GPU memory, while Llama-3.1-8B takes $217.2$s with $15.4$GB peak memory. The cost is therefore practical and amortized across downstream
  arithmetic methods.

%% file: sections/conclusion.tex
\section{Conclusion}

This work studies data-free arithmetic merging from a direction-level perspective. In an HO-GSVD-derived shared basis, task-vector matrices exhibit broadly common directions and high-dispersion unique directions whose energy is concentrated in only a subset of tasks. CUSD uses this observation to decompose task vectors before arithmetic aggregation, merge only the common components, and control the unique residuals separately. Since the decomposition depends only on model weights, the same common components and unique residuals can be reused across downstream arithmetic merge operators.

Across vision encoders, text encoders, and decoder LLMs, CUSD improves most model--operator combinations and reduces parameter-space interference. The ablations further show that effective unique-residual scaling varies across the evaluated model families and task groups, supporting the separate control of common aggregation and unique residuals. These results support separating common and task-dominated
directions before arithmetic merging.

%% file: checklist.tex
\section*{NeurIPS Paper Checklist}

\begin{enumerate}

\item {\bf Claims}
    \item[] Question: Do the main claims made in the abstract and introduction accurately reflect the paper's contributions and scope?
    \item[] Answer: \answerYes{} 
    \item[] Justification: The assertions made in the abstract and introduction are fully supported by our methodology and empirical findings.
    \item[] Guidelines:
    \begin{itemize}
        \item The answer \answerNA{} means that the abstract and introduction do not include the claims made in the paper.
        \item The abstract and/or introduction should clearly state the claims made, including the contributions made in the paper and important assumptions and limitations. A \answerNo{} or \answerNA{} answer to this question will not be perceived well by the reviewers. 
        \item The claims made should match theoretical and experimental results, and reflect how much the results can be expected to generalize to other settings. 
        \item It is fine to include aspirational goals as motivation as long as it is clear that these goals are not attained by the paper. 
    \end{itemize}

\item {\bf Limitations}
    \item[] Question: Does the paper discuss the limitations of the work performed by the authors?
    \item[] Answer: \answerYes{} 
    \item[] Justification: Please refer to Appendix~\ref{app:limitation} to see our related discussion.
    \item[] Guidelines:
    \begin{itemize}
        \item The answer \answerNA{} means that the paper has no limitation while the answer \answerNo{} means that the paper has limitations, but those are not discussed in the paper. 
        \item The authors are encouraged to create a separate ``Limitations'' section in their paper.
        \item The paper should point out any strong assumptions and how robust the results are to violations of these assumptions (e.g., independence assumptions, noiseless settings, model well-specification, asymptotic approximations only holding locally). The authors should reflect on how these assumptions might be violated in practice and what the implications would be.
        \item The authors should reflect on the scope of the claims made, e.g., if the approach was only tested on a few datasets or with a few runs. In general, empirical results often depend on implicit assumptions, which should be articulated.
        \item The authors should reflect on the factors that influence the performance of the approach. For example, a facial recognition algorithm may perform poorly when image resolution is low or images are taken in low lighting. Or a speech-to-text system might not be used reliably to provide closed captions for online lectures because it fails to handle technical jargon.
        \item The authors should discuss the computational efficiency of the proposed algorithms and how they scale with dataset size.
        \item If applicable, the authors should discuss possible limitations of their approach to address problems of privacy and fairness.
        \item While the authors might fear that complete honesty about limitations might be used by reviewers as grounds for rejection, a worse outcome might be that reviewers discover limitations that aren't acknowledged in the paper. The authors should use their best judgment and recognize that individual actions in favor of transparency play an important role in developing norms that preserve the integrity of the community. Reviewers will be specifically instructed to not penalize honesty concerning limitations.
    \end{itemize}

\item {\bf Theory assumptions and proofs}
    \item[] Question: For each theoretical result, does the paper provide the full set of assumptions and a complete (and correct) proof?
    \item[] Answer: \answerYes{} 
    \item[] Justification: Please refer to Appendix~\ref{app:lossless} to see our theoretical proof.
    \item[] Guidelines:
    \begin{itemize}
        \item The answer \answerNA{} means that the paper does not include theoretical results. 
        \item All the theorems, formulas, and proofs in the paper should be numbered and cross-referenced.
        \item All assumptions should be clearly stated or referenced in the statement of any theorems.
        \item The proofs can either appear in the main paper or the supplemental material, but if they appear in the supplemental material, the authors are encouraged to provide a short proof sketch to provide intuition. 
        \item Inversely, any informal proof provided in the core of the paper should be complemented by formal proofs provided in appendix or supplemental material.
        \item Theorems and Lemmas that the proof relies upon should be properly referenced. 
    \end{itemize}

    \item {\bf Experimental result reproducibility}
    \item[] Question: Does the paper fully disclose all the information needed to reproduce the main experimental results of the paper to the extent that it affects the main claims and/or conclusions of the paper (regardless of whether the code and data are provided or not)?
    \item[] Answer: \answerYes{} 
    \item[] Justification: We conduct detailed descriptions of our experimental setup in Section~\ref{sec:experiments}. The source code is publicly available at \url{https://github.com/stwang666/CUSD}
    \item[] Guidelines:
    \begin{itemize}
        \item The answer \answerNA{} means that the paper does not include experiments.
        \item If the paper includes experiments, a \answerNo{} answer to this question will not be perceived well by the reviewers: Making the paper reproducible is important, regardless of whether the code and data are provided or not.
        \item If the contribution is a dataset and\slash or model, the authors should describe the steps taken to make their results reproducible or verifiable. 
        \item Depending on the contribution, reproducibility can be accomplished in various ways. For example, if the contribution is a novel architecture, describing the architecture fully might suffice, or if the contribution is a specific model and empirical evaluation, it may be necessary to either make it possible for others to replicate the model with the same dataset, or provide access to the model. In general. releasing code and data is often one good way to accomplish this, but reproducibility can also be provided via detailed instructions for how to replicate the results, access to a hosted model (e.g., in the case of a large language model), releasing of a model checkpoint, or other means that are appropriate to the research performed.
        \item While NeurIPS does not require releasing code, the conference does require all submissions to provide some reasonable avenue for reproducibility, which may depend on the nature of the contribution. For example
        \begin{enumerate}
            \item If the contribution is primarily a new algorithm, the paper should make it clear how to reproduce that algorithm.
            \item If the contribution is primarily a new model architecture, the paper should describe the architecture clearly and fully.
            \item If the contribution is a new model (e.g., a large language model), then there should either be a way to access this model for reproducing the results or a way to reproduce the model (e.g., with an open-source dataset or instructions for how to construct the dataset).
            \item We recognize that reproducibility may be tricky in some cases, in which case authors are welcome to describe the particular way they provide for reproducibility. In the case of closed-source models, it may be that access to the model is limited in some way (e.g., to registered users), but it should be possible for other researchers to have some path to reproducing or verifying the results.
        \end{enumerate}
    \end{itemize}

\item {\bf Open access to data and code}
    \item[] Question: Does the paper provide open access to the data and code, with sufficient instructions to faithfully reproduce the main experimental results, as described in supplemental material?
    \item[] Answer: \answerYes{} 
    \item[] Justification: The source code is publicly available at \url{https://github.com/stwang666/CUSD}
    \item[] Guidelines:
    \begin{itemize}
        \item The answer \answerNA{} means that paper does not include experiments requiring code.
        \item Please see the NeurIPS code and data submission guidelines (\url{https://neurips.cc/public/guides/CodeSubmissionPolicy}) for more details.
        \item While we encourage the release of code and data, we understand that this might not be possible, so \answerNo{} is an acceptable answer. Papers cannot be rejected simply for not including code, unless this is central to the contribution (e.g., for a new open-source benchmark).
        \item The instructions should contain the exact command and environment needed to run to reproduce the results. See the NeurIPS code and data submission guidelines (\url{https://neurips.cc/public/guides/CodeSubmissionPolicy}) for more details.
        \item The authors should provide instructions on data access and preparation, including how to access the raw data, preprocessed data, intermediate data, and generated data, etc.
        \item The authors should provide scripts to reproduce all experimental results for the new proposed method and baselines. If only a subset of experiments are reproducible, they should state which ones are omitted from the script and why.
        \item At submission time, to preserve anonymity, the authors should release anonymized versions (if applicable).
        \item Providing as much information as possible in supplemental material (appended to the paper) is recommended, but including URLs to data and code is permitted.
    \end{itemize}

\item {\bf Experimental setting/details}
    \item[] Question: Does the paper specify all the training and test details (e.g., data splits, hyperparameters, how they were chosen, type of optimizer) necessary to understand the results?
    \item[] Answer: \answerYes{} 
    \item[] Justification: Please refer to Section~\ref{sec:experiments} for detailed empirical settings. 
    \item[] Guidelines:
    \begin{itemize}
        \item The answer \answerNA{} means that the paper does not include experiments.
        \item The experimental setting should be presented in the core of the paper to a level of detail that is necessary to appreciate the results and make sense of them.
        \item The full details can be provided either with the code, in appendix, or as supplemental material.
    \end{itemize}

\item {\bf Experiment statistical significance}
    \item[] Question: Does the paper report error bars suitably and correctly defined or other appropriate information about the statistical significance of the experiments?
    \item[] Answer: \answerNo{} 
    \item[] Justification:  The paper does not report error bars, but experiments were conducted across multiple model architectures and sizes to validate the robustness of the method.
    \item[] Guidelines:
    \begin{itemize}
        \item The answer \answerNA{} means that the paper does not include experiments.
        \item The authors should answer \answerYes{} if the results are accompanied by error bars, confidence intervals, or statistical significance tests, at least for the experiments that support the main claims of the paper.
        \item The factors of variability that the error bars are capturing should be clearly stated (for example, train/test split, initialization, random drawing of some parameter, or overall run with given experimental conditions).
        \item The method for calculating the error bars should be explained (closed form formula, call to a library function, bootstrap, etc.)
        \item The assumptions made should be given (e.g., Normally distributed errors).
        \item It should be clear whether the error bar is the standard deviation or the standard error of the mean.
        \item It is OK to report 1-sigma error bars, but one should state it. The authors should preferably report a 2-sigma error bar than state that they have a 96\% CI, if the hypothesis of Normality of errors is not verified.
        \item For asymmetric distributions, the authors should be careful not to show in tables or figures symmetric error bars that would yield results that are out of range (e.g., negative error rates).
        \item If error bars are reported in tables or plots, the authors should explain in the text how they were calculated and reference the corresponding figures or tables in the text.
    \end{itemize}

\item {\bf Experiments compute resources}
    \item[] Question: For each experiment, does the paper provide sufficient information on the computer resources (type of compute workers, memory, time of execution) needed to reproduce the experiments?
    \item[] Answer: \answerYes{} 
    \item[] Justification: We report the time consumption of our proposed method and computed resources used in the experimental setup section.
    \item[] Guidelines:
    \begin{itemize}
        \item The answer \answerNA{} means that the paper does not include experiments.
        \item The paper should indicate the type of compute workers CPU or GPU, internal cluster, or cloud provider, including relevant memory and storage.
        \item The paper should provide the amount of compute required for each of the individual experimental runs as well as estimate the total compute. 
        \item The paper should disclose whether the full research project required more compute than the experiments reported in the paper (e.g., preliminary or failed experiments that didn't make it into the paper). 
    \end{itemize}
    
\item {\bf Code of ethics}
    \item[] Question: Does the research conducted in the paper conform, in every respect, with the NeurIPS Code of Ethics \url{https://neurips.cc/public/EthicsGuidelines}?
    \item[] Answer: \answerYes{} 
    \item[] Justification: The research conducted in this paper fully conforms to the NeurIPS Code of Ethics in every respect.
    \item[] Guidelines:
    \begin{itemize}
        \item The answer \answerNA{} means that the authors have not reviewed the NeurIPS Code of Ethics.
        \item If the authors answer \answerNo, they should explain the special circumstances that require a deviation from the Code of Ethics.
        \item The authors should make sure to preserve anonymity (e.g., if there is a special consideration due to laws or regulations in their jurisdiction).
    \end{itemize}

\item {\bf Broader impacts}
    \item[] Question: Does the paper discuss both potential positive societal impacts and negative societal impacts of the work performed?
    \item[] Answer: \answerYes{} 
    \item[] Justification: We discuss both the potential positive and negative societal impacts of our
work in Appendix~\ref{app:broader_impacts}.
    \item[] Guidelines:
    \begin{itemize}
        \item The answer \answerNA{} means that there is no societal impact of the work performed.
        \item If the authors answer \answerNA{} or \answerNo, they should explain why their work has no societal impact or why the paper does not address societal impact.
        \item Examples of negative societal impacts include potential malicious or unintended uses (e.g., disinformation, generating fake profiles, surveillance), fairness considerations (e.g., deployment of technologies that could make decisions that unfairly impact specific groups), privacy considerations, and security considerations.
        \item The conference expects that many papers will be foundational research and not tied to particular applications, let alone deployments. However, if there is a direct path to any negative applications, the authors should point it out. For example, it is legitimate to point out that an improvement in the quality of generative models could be used to generate Deepfakes for disinformation. On the other hand, it is not needed to point out that a generic algorithm for optimizing neural networks could enable people to train models that generate Deepfakes faster.
        \item The authors should consider possible harms that could arise when the technology is being used as intended and functioning correctly, harms that could arise when the technology is being used as intended but gives incorrect results, and harms following from (intentional or unintentional) misuse of the technology.
        \item If there are negative societal impacts, the authors could also discuss possible mitigation strategies (e.g., gated release of models, providing defenses in addition to attacks, mechanisms for monitoring misuse, mechanisms to monitor how a system learns from feedback over time, improving the efficiency and accessibility of ML).
    \end{itemize}
    
\item {\bf Safeguards}
    \item[] Question: Does the paper describe safeguards that have been put in place for responsible release of data or models that have a high risk for misuse (e.g., pre-trained language models, image generators, or scraped datasets)?
    \item[] Answer: \answerNA{} 
    \item[] Justification: The methods primarily discussed in our paper do not involve data or models with a high risk of misuse.
    \item[] Guidelines:
    \begin{itemize}
        \item The answer \answerNA{} means that the paper poses no such risks.
        \item Released models that have a high risk for misuse or dual-use should be released with necessary safeguards to allow for controlled use of the model, for example by requiring that users adhere to usage guidelines or restrictions to access the model or implementing safety filters. 
        \item Datasets that have been scraped from the Internet could pose safety risks. The authors should describe how they avoided releasing unsafe images.
        \item We recognize that providing effective safeguards is challenging, and many papers do not require this, but we encourage authors to take this into account and make a best faith effort.
    \end{itemize}

\item {\bf Licenses for existing assets}
    \item[] Question: Are the creators or original owners of assets (e.g., code, data, models), used in the paper, properly credited and are the license and terms of use explicitly mentioned and properly respected?
    \item[] Answer: \answerYes{} 
    \item[] Justification: We have appropriately cited the papers relevant to our work.
    \item[] Guidelines:
    \begin{itemize}
        \item The answer \answerNA{} means that the paper does not use existing assets.
        \item The authors should cite the original paper that produced the code package or dataset.
        \item The authors should state which version of the asset is used and, if possible, include a URL.
        \item The name of the license (e.g., CC-BY 4.0) should be included for each asset.
        \item For scraped data from a particular source (e.g., website), the copyright and terms of service of that source should be provided.
        \item If assets are released, the license, copyright information, and terms of use in the package should be provided. For popular datasets, \url{paperswithcode.com/datasets} has curated licenses for some datasets. Their licensing guide can help determine the license of a dataset.
        \item For existing datasets that are re-packaged, both the original license and the license of the derived asset (if it has changed) should be provided.
        \item If this information is not available online, the authors are encouraged to reach out to the asset's creators.
    \end{itemize}

\item {\bf New assets}
    \item[] Question: Are new assets introduced in the paper well documented and is the documentation provided alongside the assets?
    \item[] Answer: \answerYes{} 
    \item[] Justification: The source code is publicly available at \url{https://github.com/stwang666/CUSD}
    \item[] Guidelines:
    \begin{itemize}
        \item The answer \answerNA{} means that the paper does not release new assets.
        \item Researchers should communicate the details of the dataset\slash code\slash model as part of their submissions via structured templates. This includes details about training, license, limitations, etc. 
        \item The paper should discuss whether and how consent was obtained from people whose asset is used.
        \item At submission time, remember to anonymize your assets (if applicable). You can either create an anonymized URL or include an anonymized zip file.
    \end{itemize}

\item {\bf Crowdsourcing and research with human subjects}
    \item[] Question: For crowdsourcing experiments and research with human subjects, does the paper include the full text of instructions given to participants and screenshots, if applicable, as well as details about compensation (if any)? 
    \item[] Answer: \answerNA{} 
    \item[] Justification: The paper does not involve crowdsourcing nor research with human subjects.
    \item[] Guidelines:
    \begin{itemize}
        \item The answer \answerNA{} means that the paper does not involve crowdsourcing nor research with human subjects.
        \item Including this information in the supplemental material is fine, but if the main contribution of the paper involves human subjects, then as much detail as possible should be included in the main paper. 
        \item According to the NeurIPS Code of Ethics, workers involved in data collection, curation, or other labor should be paid at least the minimum wage in the country of the data collector. 
    \end{itemize}

\item {\bf Institutional review board (IRB) approvals or equivalent for research with human subjects}
    \item[] Question: Does the paper describe potential risks incurred by study participants, whether such risks were disclosed to the subjects, and whether Institutional Review Board (IRB) approvals (or an equivalent approval/review based on the requirements of your country or institution) were obtained?
    \item[] Answer: \answerNA{} 
    \item[] Justification: The paper does not involve crowdsourcing nor research with human subjects.
    \item[] Guidelines:
    \begin{itemize}
        \item The answer \answerNA{} means that the paper does not involve crowdsourcing nor research with human subjects.
        \item Depending on the country in which research is conducted, IRB approval (or equivalent) may be required for any human subjects research. If you obtained IRB approval, you should clearly state this in the paper. 
        \item We recognize that the procedures for this may vary significantly between institutions and locations, and we expect authors to adhere to the NeurIPS Code of Ethics and the guidelines for their institution. 
        \item For initial submissions, do not include any information that would break anonymity (if applicable), such as the institution conducting the review.
    \end{itemize}

\item {\bf Declaration of LLM usage}
    \item[] Question: Does the paper describe the usage of LLMs if it is an important, original, or non-standard component of the core methods in this research? Note that if the LLM is used only for writing, editing, or formatting purposes and does \emph{not} impact the core methodology, scientific rigor, or originality of the research, declaration is not required.
    \item[] Answer: \answerNA{} 
    \item[] Justification: The core methodology of this paper does not involve LLMs as an important, original, or non-standard component. LLMs from the Llama and Gemma families appear only as experimental subjects on which the proposed merging method is evaluated. 
    \item[] Guidelines:
    \begin{itemize}
        \item The answer \answerNA{} means that the core method development in this research does not involve LLMs as any important, original, or non-standard components.
        \item Please refer to our LLM policy in the NeurIPS handbook for what should or should not be described.
    \end{itemize}

\end{enumerate}